\documentclass[letterpaper, 10 pt, journal, twoside]{IEEEtran}

\usepackage{amsmath}
\usepackage{amssymb}
\usepackage{dsfont}
\usepackage{color}
\usepackage{subcaption}
\usepackage{graphicx}
\usepackage{theorem}
\usepackage{titlesec}
\usepackage{url}
\usepackage[
colorlinks = true,
urlcolor  = magenta,
]{hyperref}
\usepackage{booktabs}
\usepackage{multirow}
\usepackage{setspace}
\usepackage{bm}
\usepackage[dvipsnames]{xcolor}

\def\CAMERAREADY{1}  

\ifnum\CAMERAREADY=1
\newcommand{\rbt}[1]{\textcolor{black}{#1}}
\newcommand{\pac}[1]{\textcolor{black}{#1}}
\newcommand{\paca}[1]{\textcolor{black}{#1}}
\else 
\newcommand{\rbt}[1]{\textcolor{teal}{#1}}
\newcommand{\pac}[1]{\textcolor{magenta}{#1}}
\newcommand{\paca}[1]{\textcolor{purple}{#1}}
\fi

\definecolor{myblue}{RGB}{0, 78, 158}
\newcommand{\cs}[1]{\textcolor{myblue}{\mathtt{#1}}}
\newcommand{\ct}[1]{\textcolor{myblue}{#1}}

\newcommand{\partitle}[1]{\noindent{\textbf{#1.}}}

\newcommand{\trajopt}[0]{\rbt{Traj. Opt.}}
\newcommand{\ournor}[0]{Ours (w/o RefineNet)}
\newcommand{\ourstl}[0]{Ours ($\mathcal{L}_{STL}$)}

\begin{document}
%
\title{Diverse Controllable Diffusion Policy with Signal Temporal Logic}
%
%
%


\author{Yue Meng$^{1}$ and Chuchu Fan$^{1}$,~\IEEEmembership{Member,~IEEE}%
\thanks{Manuscript received: March 26, 2024; Revised: June 22, 2024; Accepted: July 31, 2024.}
\thanks{
This paper was recommended for publication by Editor Jens Kober upon evaluation of the Associate Editor and Reviewers’ comments.
This work was partly supported by the National Science Foundation (NSF) CAREER Award \#CCF-2238030 and the MIT-Ford Alliance program.} 
\thanks{$^{1}$Yue Meng and Chuchu Fan are with the Department of Aeronautics and Astronautics, Massachusetts Institute of Technology, Cambridge, MA 02139 USA
{\tt\footnotesize (mengyue@mit.edu; chuchu@mit.edu)}}
\thanks{Digital Object Identifier (DOI): see top of this page.}
}

%
%

\markboth{IEEE Robotics and Automation Letters. Preprint Version. Accepted July~2024}
{Meng \MakeLowercase{\textit{et al.}}: Diverse Temporal Logic Diffusion Policy} 

%



\maketitle



%
\IEEEpeerreviewmaketitle


\begin{abstract}
Generating realistic simulations is critical for autonomous system applications such as self-driving and human-robot interactions. However, driving simulators nowadays still have difficulty in generating controllable, diverse, and rule-compliant behaviors for road participants: Rule-based models cannot produce diverse behaviors and require careful tuning, whereas learning-based methods imitate the policy from data but are not designed to follow the rules explicitly. Besides, the real-world datasets are by nature ``single-outcome", making the learning method hard to generate diverse behaviors. In this paper, we leverage Signal Temporal Logic (STL) and \rbt{Diffusion Models} to learn controllable, diverse, and rule-aware policy. We first calibrate the STL on the real-world data, then generate diverse synthetic data using trajectory optimization, and finally learn the rectified diffusion policy on the augmented dataset. We test on the NuScenes dataset and our approach can achieve the most diverse rule-compliant trajectories compared to other baselines, with a runtime 1/17X to the second-best approach. In the closed-loop testing, our approach reaches the highest diversity, rule satisfaction rate, and the least collision rate. Our method can generate varied characteristics conditional on different STL parameters in testing. \rbt{A case study on human-robot encounter scenarios shows our approach can generate diverse and closed-to-oracle trajectories.} The annotation tool, \pac{augmented dataset, and code are available} at \url{https://github.com/mengyuest/pSTL-diffusion-policy}.
\end{abstract}

\begin{IEEEkeywords}
Autonomous Agents; Autonomous Vehicle Navigation; Machine Learning for Robot Control
\end{IEEEkeywords}
\section{Introduction}
\label{sec:intro}
\IEEEPARstart{R}{ealistic} behavior modeling is vital for developing simulators and studying intelligent systems such as autonomous driving and warehouse ground robots~\cite{jager2002need,suo2021trafficsim}. To close the sim-to-real gap for the agents, it is critical to model the uncertainty and rule adherence properties that naturally arise from human behaviors. For example, human drivers have different characteristics (aggressiveness, conservativeness), which affect their decision-making in challenging scenarios (e.g., going through a roundabout with dense traffic). Besides, low-level driving commands (steering the wheel, accelerating, braking) are also driven by high-level maneuvers (lane-keeping, lane-changing) and traffic rules (e.g., speed limit). Thus, it is of paramount importance to endow agent models with diversity, controllability, and rule-awareness.

However, driving simulators up-to-date~\cite{dosovitskiy2017carla,lopez2018microscopic} still struggle in delivering diverse and rule-compliance agent behaviors. They either use recorded trajectories or utilize rule-based or imitation-based methods to generate policy. Rule-based approaches (IDM~\cite{kesting2010enhanced}, MOBIL~\cite{kesting2007general}) directly encode rules into mathematical models thus can provide safety and goal-reaching performance. Still, they assume simplified driving scenarios and require careful parameter tuning, lacking diversity and realisticness. Imitation-based methods~\cite{salzmann2020trajectron++,bhattacharyya2022modeling} learn from real-world driving data, being more akin to human behaviors, but are prone to violate the rules. Besides, since there is only one outcome (out of many possible future trajectories) per scene in the ground truth, only a limited diversity is achieved by these imitation-based approaches~\cite{liang2020garden}.

Impeding the advancement of learning realistic behaviors are three challenges: (1) A flexible rule representation, (2) the scarcity of multiple-outcome datasets, and (3) the trade-off between rule compliance and diversity. Our paper systematically addresses these problems by leveraging a formal language termed Signal Temporal Logic (STL)~\cite{donze2010robust,donze2013efficient}. STL is known for modeling complicated rules~\cite{sahin2020autonomous}, and there are increasing works recently studying controller synthesis under STL specifications via trajectory optimization~\cite{dawson2022robust}, deep learning~\cite{leung2023backpropagation,meng2023signal} and reinforcement learning~\cite{li2017reinforcement}. Inspired by these works and recent breakthroughs in diffusion models~\cite{ho2020denoising} for policy learning~\cite{zhong2023guided}, we proposed a parametric-STL approach to flexibly encode traffic rules, augment the dataset, and learn a controllable diffusion policy to balance \pac{quality and diversity}\footnote{\rbt{``Diversity" refers to generate \textbf{different} trajectories for the \textbf{same} STL rule.}}.

The whole pipeline is: \rbt{We first} specify rules via parameter-STL and use demonstrations to calibrate the parameters. The parameters involve both discrete and continuous values, adding the capacity to form multi-modal and diverse policy distributions. Based on the STL, the parameters, and the original data, we generate the ``multiple-outcome" data via trajectory optimization. Next, we use \rbt{Denoising Diffusion Probabilistic Model (DDPM~\cite{ho2020denoising}) to learn from the augmented data.} \pac{Finally, different from other diffusion-based policies~\cite{zhong2023guided,chi2023diffusion}, } {an additional neural network is designed to regulate the trajectories to be rule-compliant and diverse.}

We conduct experiments on NuScenes~\cite{caesar2020nuscenes}, a large-scale autonomous driving dataset. We first label the dataset using our annotation tool and generate the augmented dataset. Then we train our approach on the augmented dataset and evaluate on the validation set and \rbt{in} closed-loop testing. Our approach results in the highest STL satisfaction rate on the validation set and generates the most diverse trajectories compared to baselines. In closed-loop testing, our approach reaches the highest overall performance regarding diversity, STL satisfaction, collision, out-of-lane, and progress. We also show that \pac{with varied STL parameters, our} approach can reflect different driver characteristics in a challenging roundabout scenario, which is valuable for diverse behavior modeling in simulators. \rbt{A case study on human-robot encounter scenarios also shows similar performance compared to other baselines.}

\label{sec:contribution}
To summarize, our contributions are\rbt{: (1) We are the first to use a parametric-STL formulation to augment the driving dataset for diverse and controllable policy generation (2) we propose an add-on module (RefineNet) for Diffusion Models to \pac{improve trajectory diversity and quality} (3) we achieve leading performance in the open-loop evaluation and closed-loop test on NuScenes~\cite{caesar2020nuscenes} (4) our algorithm, annotation tool and the augmented data will be available via open-source distribution.}
\section{Related Work}
\label{sec:related}
\partitle{Trajectory prediction}
Decades of effort have been devoted to exploring trajectory prediction for autonomous systems~\cite{leon2021review}. Traditional methods include physics-based methods~\cite{ammoun2009real} and machine-learning approaches such as Gaussian Process~\cite{joseph2011bayesian}. More recent works use neural networks to conduct behavior cloning or imitation learning~\cite{ly2020learning} on multi-modal large-scale datasets (NuScenes~\cite{caesar2020nuscenes}, WOMD~\cite{sun2020scalability}), where the performance is improved by better scene representations (rasterization~\cite{salzmann2020trajectron++}, polylines~\cite{gao2020vectornet}), advanced architectures~\cite{gupta2018social}, and varied output types (sets~\cite{phan2020covernet}, heatmaps~\cite{gilles2021thomas} and distributions~\cite{jiang2023motiondiffuser}). Most works predict trajectories by fitting the dataset. Instead, we assume the rules are given, extract statistics from the data and learn diverse and rule-compliant trajectories conditional on these statistics.

\partitle{Diverse trajectory generation} Deep generative models (Variational Auto Encoder, Generative Adversarial Network, and \rbt{Diffusion Models}) are used to learn to produce diverse trajectories\paca{~\cite{gupta2018social,zhong2023guided,chi2023diffusion,reuss2023goal,scheikl2024movement}}. The diversity is either learned implicitly from data\paca{~\cite{zhong2023guided,chi2023diffusion,reuss2023goal,scheikl2024movement}} or guided by the ``Minimum over N" (MoN) loss~\cite{gupta2018social}. To avoid the prediction concentrating merely around the major mode of the data, the work in~\cite{yuan2019diverse} learns diversity sampling functions (DSF) in the latent space to generate diverse trajectories. In the autonomous driving domain, recent works use inductive heuristics from the scenes (such as drivable area~\cite{xu2024controllable} or lanes~\cite{kim2022diverse}) to further regulate the diverse behaviors to be ``reasonable" in common sense. However, it is worth noting that the original dataset lacks diversity in essence (for each scene, there is just one ground trajectory). Unlike all methods that learn diversity from the original data, we first generate diverse data using trajectory optimization then learn the diverse policy. The closest works similar to ours are ForkingPath~\cite{liang2020garden} and \cite{stoll2023scaling}, where the former is for pedestrian prediction and requires heavy annotation, and the latter generates data using IDM  which are less diverse.

\partitle{Realistic agent modeling} Realisticness is often achieved by augmenting imitation with common sense factors, such as collision-free~\cite{suo2021trafficsim,meng2021reactive}, map-consistency~\cite{casas2020importance}, attractor-repeller effect~\cite{jiang2023motiondiffuser}, driving patterns~\cite{bhattacharyya2019simulating}, and LLM-based designs~\cite{zhong2023language}. Recently, Signal Temporal Logic (STL)~\cite{donze2010robust} is widely used to specify rules for trajectory predictions~\cite{maierhofer2022formalization,zhong2023guided}, for its expressiveness to encode rules~\cite{maierhofer2022formalization} and differentiable policy learning~\cite{leung2023backpropagation,meng2023signal}. We follow this line of work and parametrize the STL to \pac{learn controllable behaviors}.
\section{Preliminaries}
\label{sec:prelim}
\subsection{Signal Temporal Logic (STL)}
\rbt{A signal $s=x_t,x_{t+1},...,x_{t+T}$ is a discrete-time finite sequence of states $x_i\in\mathbb{R}^n$. STL is a formal language to specify signal properties via the following expressions~\cite{donze2013efficient}:}
\begin{equation}
    \phi::= \top \ | \ \mu(x)\geq 0 \ | \ \neg \phi \ | \ \phi_1 \land \phi_2 \ | \ \phi_1 \text{U}_{[a,b]}\phi_2
    \label{eq:stl_formula}
\end{equation}
\rbt{where the boolean-type (``true" / ``false") operators split by ``$|$" serve as building blocks to construct an STL formula. Here} $\top$ is ``true", $\mu$ \rbt{denotes a function} $\mathbb{R}^n\to\mathbb{R}$, and $\neg$, $\land$, $\text{U}$, ${[a,b]}$ are ``not", ``and", ``until'', and \rbt{time interval from $a$ to $b$}. 
Other operators are ``or": $\phi_1\lor\phi_2=\neg(\neg\phi_1 \land \neg\phi_2)$, ``imply": $\phi_1\Rightarrow\phi_2=\neg\phi_1 \lor \phi_2$, ``eventually": $\lozenge_{[a,b]} \phi=\top \text{U}_{[a,b]}\phi$ and ``always": $\square_{[a,b]} \phi = \neg\lozenge_{[a,b]}\neg\phi$. Denote $s,t\models \phi$ if a signal $s$ from time $t$ \rbt{satisfies $\phi$, i.e., $\phi$ returns ``true". It is easy to check satisfaction for $\top$, $\mu\geq 0$,  $\neg$, $\land$, and $\lor$.} As for temporal operators~\cite{maler2004monitoring}: $s,t \models \lozenge_{[a,b]}\phi  \Leftrightarrow \,\, \exists t' \in [t+a, t+b]\,\, s,t'\models \phi$ and $s,t \models \square_{[a,b]}\phi  \Leftrightarrow \,\, \forall t' \in [t+a, t+b]\,\, s,t'\models \phi$.

Robustness score $\rho(s,t,\phi)$ measures how well a signal $s$ satisfies $\phi$\rbt{, where} $\rho\geq 0$ if and only if $s,t\models \phi$. A larger $\rho$ reflects a greater satisfaction margin. The calculation is~\cite{donze2010robust}:
\begin{equation}
    \begin{aligned}
        & \rho(s,t, \top)= 1,\quad  \rho(s,t, \mu)= \mu(s(t)) \\ &\rho(s,t, \neg \phi)= -\rho(s,t,\phi)\\
        & \rho(s,t, \phi_1 \land \phi_2)= \min\{\rho(s,t,\phi_1),\rho(s,t,\phi_2) \}\\
        & \rho(s,t, \lozenge_{[a,b]} \phi)= \sup\limits_{r\in[a,b]}\rho(s,t+r,\phi)\\
        & \rho(s,t, \square_{[a,b]} \phi)= \inf\limits_{r\in[a,b]}\rho(s,t+r,\phi)
    \end{aligned}
    \label{eq:robustness_score}
\end{equation}
\rbt{In our paper, we adopt a differentiable approximation for $\rho$ proposed in~\cite{pant2017smooth} to provide gradient-based policy guidance.}

\subsection{Denoising Diffusion Probabilistic Models (DDPM)}

\rbt{Diffusion Models are powerful generative models that learn a density distribution from the training data to generate samples that resemble these data. A diffusion model consists of two procedures: forward process (diffusion) and inverse process (denoising). In paper~\cite{ho2020denoising}, during the diffusion process, the data are iteratively fused with Gaussian noise until they are close to the white noise. A neural network is trained to predict the noises added to these samples at different diffusion steps. To generate samples, in the denoising process, the latent samples initialized from the white noise are recovered iteratively by ``removing" the noise predicted by the network.}

\begin{figure*}[!tbp]
\centering
    \includegraphics[width=1.0\textwidth]{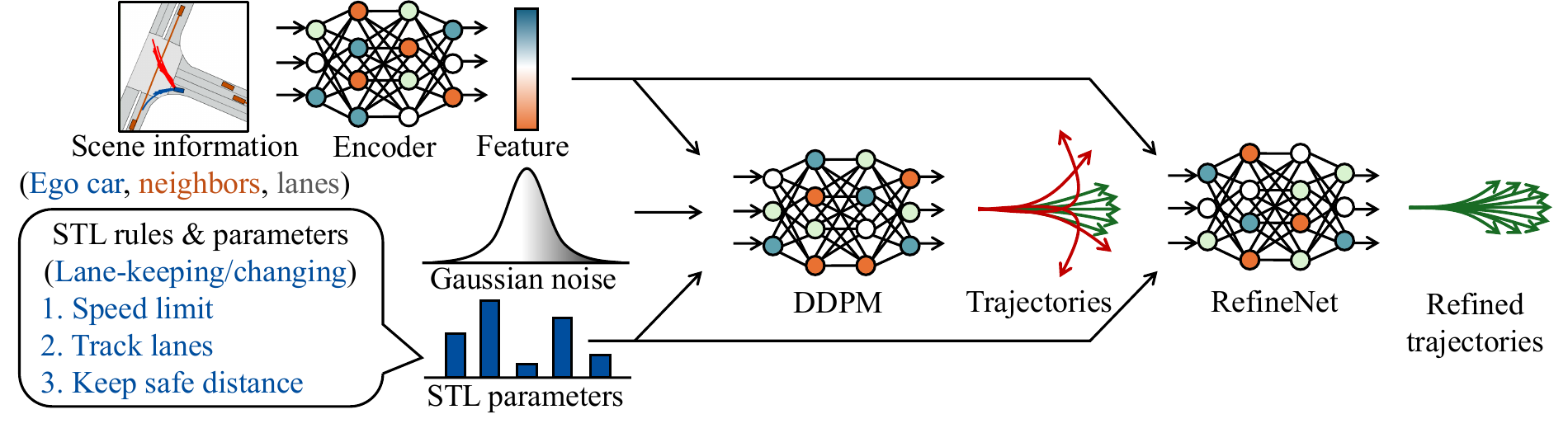}
    \caption{Learning framework. The neural encoder embeds the scene to a feature vector. The DDPM takes the feature vector, the STL parameters (indicating driving modes, speed limit, safe distance threshold, etc) and the Gaussian noise to generate trajectories. RefineNet takes the upstream trajectories and features and generates diverse and rule-compliant trajectories.}
    \label{fig:arch}
\end{figure*}

\section{Technical Approach}
\label{sec:tech}

\subsection{Problem formulation}
Consider an autonomous system, where we denote the state of the agent at time t as $s_t\in \mathcal{S}\subseteq\mathbb{R}^n$, the control $u_t\in\mathcal{U}\subseteq\mathbb{R}^m$, the scene context $c \in \mathcal{C}\subseteq\mathbb{R}^p$, and the STL template $\Psi:\Gamma\times \mathcal{C} \to \Phi$ which generates the STL formula $\phi\in\Phi$ based on the STL parameters $\gamma\in\Gamma$ and the scene provided (i.e., $\phi=\Psi(\gamma, c)$). Assume a known differentiable discrete-time system dynamics: $f:\mathcal{S}\times \mathcal{U}\to \mathcal{S}$, where from $s_0$ we could generate a trajectory $\tau=(s_0, s_1, ..., s_T) \in \mathbb{R}^{(T+1)\times n}$ based on control sequence $(u_0,u_1,...,u_{T-1})\in\mathcal{U}^T$. Assume a known diversity measure $J:\Xi\to\mathbb{R}$ which is a real-valued function over a set of trajectories $\Xi$. 

Given a set of demonstrations $\mathcal{D}=\{(c^i, \tau^i\}_{i=1}^N$ and an STL template $\Psi$, the goal is first to generate a set of STL parameters $\{\gamma_i\}_{i=1}^N$ such that $\tau_i,0 \models \Psi(\gamma_i, c_i)$, $\forall i=1,2,...N$, and secondly, learn a policy $\pi:\mathcal{S}\times \mathcal{C} \times\Gamma\to\mathcal{U}^T$ such that for a given initial state $s_0$, \rbt{a} scene context state $c$ and STL parameters $\gamma$, the trajectories set $\Xi$ generated by $\pi(s_0,c,\gamma)$ can both (1) satisfy the STL rule $\tau,0 \models \Psi(\gamma, c)$, $\forall \tau \in \Xi$, and (2) maximize the diversity measure \rbt{$J(\Xi)$} (we will use entropy to measure the diversity and the detailed computation for the entropy is shown in Sec.~\ref{sec:exp-open-loop}).

\subsection{System modeling and STL rules for autonomous driving}
\label{sec:sys-model}
\partitle{Ego car model} We use a unicycle model for the dynamics: $x_{t+1}=x_t+v_t\cos(\theta_t)\Delta t$,  $y_{t+1}=y_t+v_t\sin(\theta_t)\Delta t$, $\theta_{t+1}=\theta_t+w_t\Delta t$, $v_{t+1}=v_t+a_t\Delta t$
where the state $s_t=(x_t,y_t,\theta_t,v_t)^T$ stands for ego car's xy coordinates, heading angle, and the velocity at time $t$, and the controls $u_t=(w_t, a_t)^T$ are angular velocity and acceleration. We denote the ego car width $W$ and length $L$. 

\partitle{Scene model} The scene context consists of neighbor vehicles and lanes. We consider the $N_n$ nearest neighbors within the $R$-radius disk centered at the ego vehicle and denote their states at time $t$ as $\mathcal{N}=\{I^j_t, x^j_t, y^j_t, \theta^j_t,v^j_t,L^j,W^j\}_{j=1}^{N_n}$, where the binary indicator $I^j_t$ is one when the $j$\textsuperscript{th} neighbor is valid and zero otherwise (this happens when there are less than $N_n$ neighbor vehicles within the $R$-radius disk), and the rests are similar to the ego car state. 
\rbt{We represent each lane as $Q=(I, x_1,y_1,\theta_1,...,x_{N_p},y_{N_p},\theta_{N_p})\in\mathbb{R}^{3N_p+1}$ with an indicator to show its validity and then a sequence of $N_p$ centerline waypoints with 2D coordinates and directions. We denote the current lane, the left adjacent lane, and the right adjacent lane for the ego vehicle as $Q_c$, $Q_l$, and $Q_r$. }

\partitle{STL rules and parameters} 
We define common STL rules that are required in autonomous driving scenarios\rbt{,}
\begin{equation}
\begin{aligned}
    &\phi_0=\square_{[0,T]} \, \cs{v_{min}}\leq \text{Speed}(s) \leq \cs{v_{max}}\\
    &\phi_1=\square_{[0,T]}\, \text{Dist}(s,\mathcal{N}) \geq \cs{d_{safe}}\\
    &\phi_2=\square_{[0,T]}\, \cs{d_{min}} \leq \text{Dist}(s, Q_c) \leq \cs{d_{max}}\\
    &\phi_3=\square_{[0,T]}\, |\text{Angle}(s, Q_c)| \leq \cs{\theta_{max}}\\
    &\phi_4^{H}=\lozenge_{[0,T]}\square_{[0,T]} \cs{d_{min}} \leq \text{Dist}(s, Q_H) \leq \cs{d_{max}}\\
    &\phi_5^{H}=\lozenge_{[0,T]}\square_{[0,T]} |\text{Angle}(s, Q_H)| \leq \cs{\theta_{max}}
\end{aligned}
\end{equation}
where the \ct{variables in blue} are STL parameters, $\phi_0$ is for the speed limit, $\phi_1$ specifies the distance to the neighbors should always be greater than the threshold. Formulas $\phi_2$ and $\phi_3$ restrict the vehicle's distance and heading deviation from the current lane, whereas $\phi_4^H$ and $\phi_5^H$ are for the left and right adjacent lanes with $H\in\{l,r\}$ to restrict the car to eventually keep the distance/angle deviation from the target lane within the limit \footnote{\rbt{Since we aim to eventually ``always" keep it in the limit, the "Always" operator is put inside the "Eventually" scope to realize this behavior.}}. We consider the driving mode $\mathcal{M}$ belongs to one of the high-level behaviors: lane-keeping ($\mathcal{M}=0$), left-lane-change ($\mathcal{M}=1$) and right-lane-change($\mathcal{M}=2$). Denote the STL parameters $\gamma=(\mathcal{M}, v_{min}, v_{max}, d_{safe}, d_{min}, d_{max}, \theta_{max})^T\in \Gamma\subseteq\mathbb{R}^7$. The STL template thus is defined as:
\begin{equation}
\begin{aligned}
    \Psi(\gamma)= \left((\mathcal{M}=0)\Rightarrow (\phi_0 \land \phi_1 \land \phi_2 \land \phi_3)\right) \\ 
    \land \left((\mathcal{M}=1)\Rightarrow (\phi_0 \land \phi_1 \land \phi_4^l \land \phi_5^l)\right)\\
    \land \left((\mathcal{M}=2)\Rightarrow (\phi_0 \land \phi_1 \land \phi_4^r \land \phi_5^r)\right)
\end{aligned}
\end{equation}

\subsection{STL parameters calibration}
\label{sec:calibration}
Given the STL template $\Psi$ and the expert demonstrations $\mathcal{D}$, we find STL parameters $\gamma_i$ for each trajectory $\tau_i$ and the scene $c_i$ such that $\tau_i,0\models \Psi(\gamma_i,c_i)$. We develop an annotation tool to manually label the high-level behaviors $\mathcal{M}$ for all the trajectories, then based on $\mathcal{M}$, we calibrate the rest of the STL parameters by making the robustness score on the expert trajectories equal to zero. This is done by conducting min/max extraction on the existing measurements: e.g., to find $d_{min}$, we first check the high-level policy, then compute the minimum distance from the trajectory $\tau$ to the target lane.

\subsection{Diverse data augementation}
After obtaining the STL parameters for each scene, we augment the original demonstration by generating diverse behaviors using trajectory optimization. For each scene $c_i$ and parameter $\gamma_i$, we formulate the following optimization:
\begin{equation}
\begin{aligned}
    \mathop{\text{minimize}}\limits_{u_0,u_1,...,u_{T-1}} \quad & \sigma_+(-\rho((s_0,s_1,...,s_T),0,\Psi(\gamma_i,c_i)))\\
    \text{subject to} \quad &  u_{min}\preccurlyeq u_t \preccurlyeq u_{max}, \forall t=0,1,...,T-1\\
    &s_{t+1}=f(s_t,u_t), \forall t=0,1,...,T-1\\
\end{aligned}
\label{eq:trajopt}
\end{equation}
where $\sigma_+(\cdot)=\max(\cdot,0)$, $a \preccurlyeq b$ means the vector $a$ is elementwise no larger than the vector $b$, and $u_{min}$, $u_{max}$ are predefined control limits. The system dynamics and STL formula make Eq~\eqref{eq:trajopt} a nonlinear optimization, and we use a gradient-based method to solve for the solution. \rbt{To increase solutions' diversity, we consider all three driving modes for $\mathcal{M}$, and for each mode, we run gradient-descent from $K$ initial solutions uniformly sampled from the solution space $\mathcal{U}^T$.} 
We denote our augmented dataset as $\tilde{\mathcal{D}}=\{(\tilde{c}_i, \tilde{\gamma}_i, \{\tilde{\tau}_i^j\}_{j=1}^K)\}_{i=1}^{3N}$ where \rbt{$\tilde{c}_i=c_{\lfloor i/3 \rfloor}$ and $\tilde{\gamma}_i=\gamma_{\lfloor i/3 \rfloor}$. Here, $\lfloor \cdot \rfloor$ denotes rounding a float number to an integer index}.

\subsection{Policy learning framework}
\label{sec:method-e}
Given the new dataset $\tilde{\mathcal{D}}$, we learn the diverse and rule-compliant behavior via the learning framework shown in Fig.~\ref{fig:arch}. The encoder network embeds the ego state and the scene to a feature vector. The DDPM network takes the feature vector, STL parameters, and a Gaussian noise to produce trajectories that closely match the distribution in $\tilde{\mathcal{D}}$. Finally, the RefineNet takes upstream features and trajectories to generate diverse and rule-compliant trajectories.

\partitle{Encoder Network} The ego state and the scene are first transformed to the ego frame, $\tilde{s}$ and $\tilde{c}=\{\tilde{Q}_c,\tilde{Q}_l,\tilde{Q}_r,\tilde{\mathcal{N}}\}$ accordingly. The state $\tilde{s}$ is sent to a fully connected network (FCN) to get ego feature: $z_{ego}=g_{ego}(\tilde{s})\in\mathbb{R}^{d}$. Similarly, the lanes are fed to a lane FCN to generate feature: $z_{lane}=[g_{lane}(\tilde{Q}_c),g_{lane}(\tilde{Q}_l),g_{lane}(\tilde{Q}_r))]\in\mathbb{R}^{\rbt{3 d}}$, where $[\cdot,\cdot...]$ is the vector concatenation. To make the neighbors feature not depend on the neighbor orders, we utilize permutation-invariant operators in~\cite{qi2017pointnet} with a neighbor FCN $g_{nei}$ to get: $z_{nei}=[\max\limits_{j}g_{nei}(\tilde{\mathcal{N}}_j),\min\limits_{j}g_{nei}(\tilde{\mathcal{N}}_j),\sum\limits_{j}g_{nei}(\tilde{\mathcal{N}}_j)]\in\mathbb{R}^{\rbt{3 d}}$, where $\tilde{\mathcal{N}}_j$ is the j\textsuperscript{th} neighbor feature. The final \rbt{merged embedding is:} $z=[z_{ego}, z_{lane}, z_{nei}]\in\mathbb{R}^{\rbt{7 d}}$.

\partitle{\rbt{DDPM Network}} Given the embedding $z$, the STL parameters $\gamma$, the sample $\tau^{(0)}$ from $\tilde{\mathcal{D}}$, the random noise $\epsilon$, we generate the diffused samples: $\tau^{(t)}=\sqrt{\bar{\alpha_t}} \tau^{(0)}+\sqrt{1-\bar{\alpha_t}} \epsilon$ for uniformly sampled diffusion steps $t\sim \text{Uniform}(1,T_d)$ \rbt{and pre-defined coefficients $\alpha_t$ and $\bar{\alpha}_t=\prod\limits_{s=1}^t \alpha_s$}. The \rbt{DDPM network} $g_{d}:\mathbb{R}^{7d + 7 + 2T+  1}\to\mathbb{R}^{T\times 2}$ takes $z,\gamma,\tau^{(t)},t$ as input and predicts the noise, guided by the diffusion loss in the first stage of the training:
\begin{equation}
    \begin{aligned}
\mathcal{L}_{d}=\mathbb{E}_{\tilde{\mathcal{D}},t}\left[\left|\epsilon-g_d(z,\gamma,\tau^{(t)},t)\right|^2\right]
    \end{aligned}
    \label{eq:diffusion-loss}
\end{equation}
In inference, from the Gaussian noise $\tau^{(T_d)}\sim \mathcal{N}(0,I)$, the trajectories are generated iteratively by the denoising step: $\tau^{(t-1)}=\frac{1}{\sqrt{\alpha_t}} \left(\tau^{(t)}-\frac{1-\alpha_t}{\sqrt{1-\bar{\alpha}_t}}g_d(z, \gamma, \tau^{(t)}, t)\right) + \sigma_t \xi_t$ with $\xi_t\sim\mathcal{N}(0, I)$ and $\sigma_1=0$ and $\sigma_t=1$ for $t\geq 2$. We denote the trajectories generated by DDPM as $\tau_d$.

\partitle{Refine Network}
After DDPM is trained, we use RefineNet, a fully-connected network \rbt{$g_r:\mathbb{R}^{7d+7+2T} \to \mathbb{R}^{T\times 2}$}, to regulate the trajectories generated by the \rbt{DDPM network} to encourage rule-compliance and diversity. RefineNet \rbt{takes as input the trajectories with the highest STL score from the last five denoising steps and} outputs a residual control sequence conditional on the violation of the \rbt{STL rules, which is:}
\begin{equation}
    \begin{aligned}
        \tau_{final} = \tau_{d} + \mathds{1}\{\rho(\tau_d, 0,\Psi(\gamma, c))<0\} \cdot g_r(z,\gamma,\tau_d)
    \end{aligned}
\end{equation}
If the DDPM \rbt{produced} trajectories already satisfy the STL rules, the RefineNet will \rbt{not affect} the final \rbt{trajectories} (i.e., $\tau_{final}=\tau_d$); otherwise, the RefineNet \rbt{improves trajectories' diversity and the rule satisfaction rate.}\footnote{\rbt{Table~\ref{tab:open-loop} shows that DDPM results in a very low rule satisfaction rate, which reflects the great potential of using RefineNet for improvement.}} \pac{It is hard to directly optimize for the diversity measure (entropy approximation requires state space discretization, which is non-differentiable).} Instead, in the second stage, the RefineNet is updated by the following loss:
\begin{equation}
    \begin{aligned}
        \mathcal{L}_{r}=\mathbb{E}_{\tilde{\mathcal{D}}} \left[
        \text{tr}\left(I-(\mathcal{K}(\{\tau_{final,j}\}_{j=1}^{N_d})+I)^{-1}\right)
        \right]
    \end{aligned}
    \label{eq:diversity-loss}
\end{equation}
where $\text{tr}(\cdot)$ is the matrix trace, and $\mathcal{K}\in\mathbb{R}^{N_d\times N_d}$ is the Direct Point Process (DPP) kernel~\cite{yuan2019diverse} over $N_d$ samples:
\begin{equation}
    \begin{aligned}
        \mathcal{K}_{ij}=\mathds{1}(\rho(\tau_i)\geq 0) \cdot \exp(-|\tau_i-\tau_j|^2)\cdot \mathds{1}(\rho(\tau_j)\geq 0).
    \end{aligned}
    \label{eq:kernel}
\end{equation}
\rbt{Minimizing Eq.~\eqref{eq:diversity-loss} increases the trajectory cardinality and quality~\cite{yuan2019diverse}, thus increases the diversity and rule satisfaction.}

\subsection{Guidance-based online policy refinement}
\label{sec:method-f}
In evaluation, our learned policy might violate the STL rules in the unseen scenarios due to the generalization error. Similar to~\cite{zhong2023guided}, we use the STL guidance to improve the sampling process. For a new state $s_0$, scene $c$, the embedding $z$ and parameter $\gamma$ in testing, we replace the original denoising step to $\tau^{(t-1)}=\frac{1}{\sqrt{\alpha_t}}\tilde{\tau}^{(t)}+\sigma_t\xi_t$, where $\tilde{\tau}^{(t)}$ is initialized as $\tau^{(t)}-\frac{1-\alpha_t}{\sqrt{1-\bar{\alpha}_t}}g_d(z, \gamma, \tau^{(t)}, t)$ and is updated by minimizing $-\rho(\tilde{\tau}^{(t)}, 0, \Psi(\gamma, c))$ via gradient descent. The work~\cite{zhong2023guided} conducts multiple guidance steps at every denoising step. In contrast, we find it sufficient to just conduct the guidance step at the last several denoising steps in the diffusion model to accelerate the computation.
Although nonlinear optimization \rbt{does not} guarantee optimality, in practice, this method can satisfy the rules with high probability, as shown below.

\section{Experiments}
\label{sec:exp}
\begin{table*}[!htbp]
\centering
\caption{Open-loop evaluation: The highest is shown in \textbf{bold} and the second best is shown in \underline{underline}. Our data augmentation boosts the diversity for baselines VAE and DDPM. ``Ours+guidance" generates the trajectories in the highest quality (Success and Compliance) and diversity (Valid area, Entropy), with the runtime 1/17X to the second best CTG~\cite{zhong2023guided}.}
\begin{tabular}{ccccccc}
\toprule
Methods & Augmentation & Success $\uparrow$ & Compliance $\uparrow$ & Valid area $\uparrow$ &  Entropy $\uparrow$ 
 &Time (s) $\downarrow$  \\ 
\midrule
\textit{\trajopt{}} & Yes & \textit{0.961} & \textit{0.746} & \textit{49.13\rbt{0}} & \textit{2.124} & \textit{\rbt{36.205}} \\
VAE & - & 0.337 & 0.077 & 0.618 & 0.162 & \textbf{0.036} \\
VAE & Yes & 0.253 & 0.018 & 0.627 & 0.2\rbt{00} & \underline{0.039} \\
DDPM~\cite{ho2020denoising} & - & 0.514 & 0.078 & 1.760 & 0.455 & 0.081 \\
DDPM~\cite{ho2020denoising} & Yes & 0.548 & 0.050 & 3.444 & 0.557 & 0.081 \\
TrafficSim~\cite{suo2021trafficsim} & Yes & 0.699 & 0.335 & 6.798 & 1.059 & 0.037 \\
CTG~\cite{zhong2023guided} & Yes & \underline{0.833} & 0.267 & 14.933 & 1.384 & 13.582 \\
\midrule
\rbt{\ournor} & \rbt{Yes} & \rbt{0.624} & \rbt{0.078} & \rbt{5.899} & \rbt{0.780} & \rbt{0.172}\\
\rbt{\ourstl} & \rbt{Yes} & \rbt{0.817} & \rbt{\textbf{0.573}} & \rbt{17.032} & \rbt{1.152} & \rbt{0.174}\\
Ours & Yes & 0.782 & 0.442 & \underline{20.284} & \underline{1.411} & 0.174 \\
Ours+guidance & Yes & \textbf{0.840} & \underline{0.544} & \textbf{33.530} & \textbf{1.735} & 0.786 \\
\bottomrule
\end{tabular}
\label{tab:open-loop}
\end{table*}
\begin{figure*}[!htbp]
    \centering
    \begin{subfigure}[b]{0.15\textwidth}
        \centering
        \includegraphics[width=\textwidth]{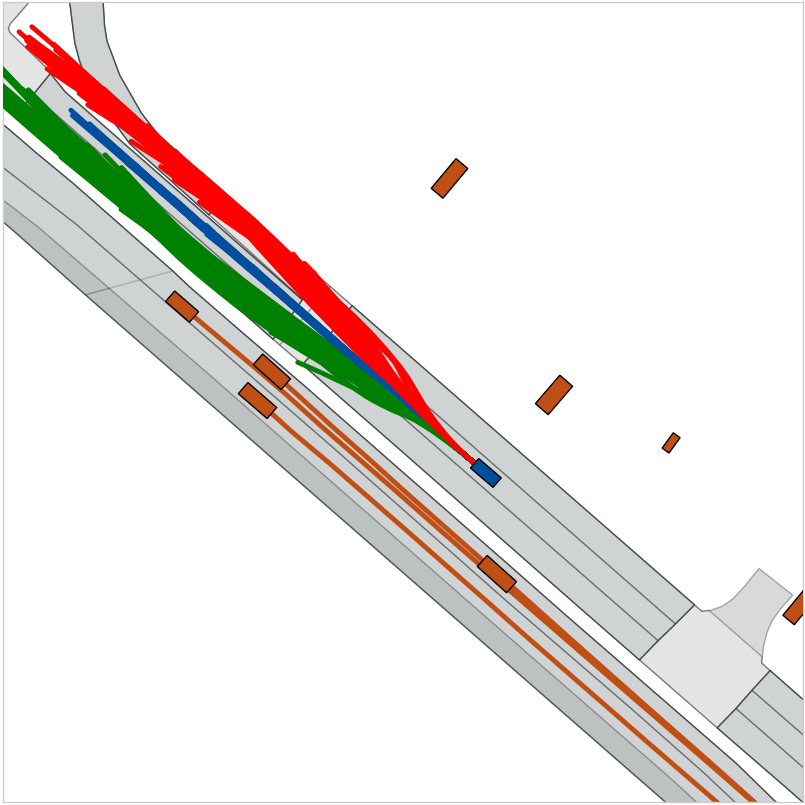}
        \caption{\textit{\trajopt{}}}
        \label{fig:bc}
    \end{subfigure}
    \begin{subfigure}[b]{0.15\textwidth}
        \centering
        \includegraphics[width=\textwidth]{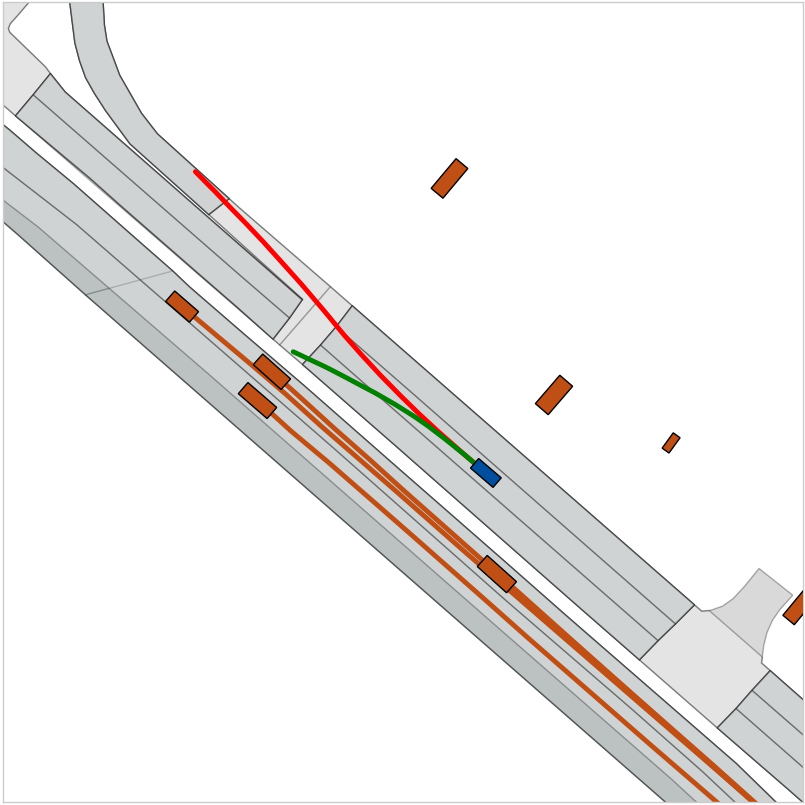}
        \caption{VAE}
        \label{fig:vae0-stl}
    \end{subfigure}
    \begin{subfigure}[b]{0.15\textwidth}
        \centering
        \includegraphics[width=\textwidth]{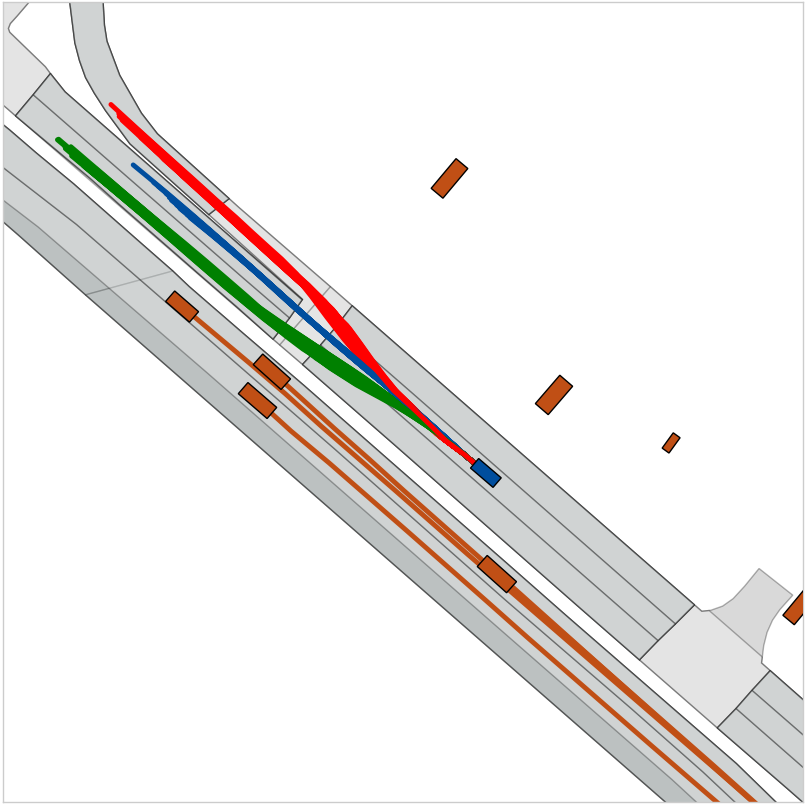}
        \caption{TrafficSim}
        \label{fig:bc-stl}
    \end{subfigure}
    \begin{subfigure}[b]{0.15\textwidth}
        \centering
        \includegraphics[width=\textwidth]{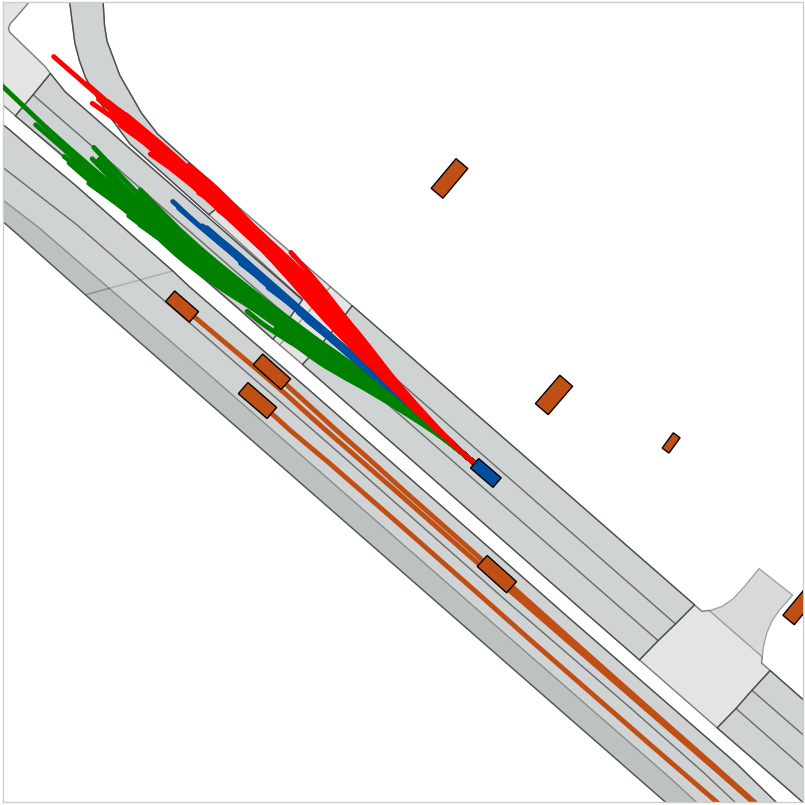}
        \caption{CTG}
        \label{fig:vae}
    \end{subfigure}
    \begin{subfigure}[b]{0.15\textwidth}
        \centering
        \includegraphics[width=\textwidth]{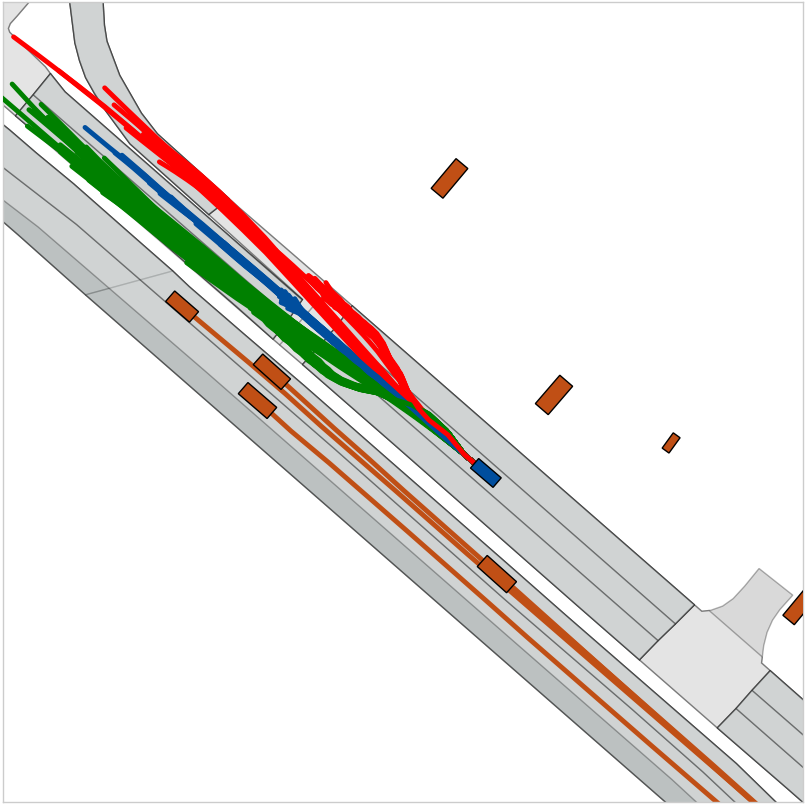}
        \caption{Ours}
        \label{fig:vae-stl}
    \end{subfigure}
    \begin{subfigure}[b]{0.15\textwidth}
        \centering
        \includegraphics[width=\textwidth]{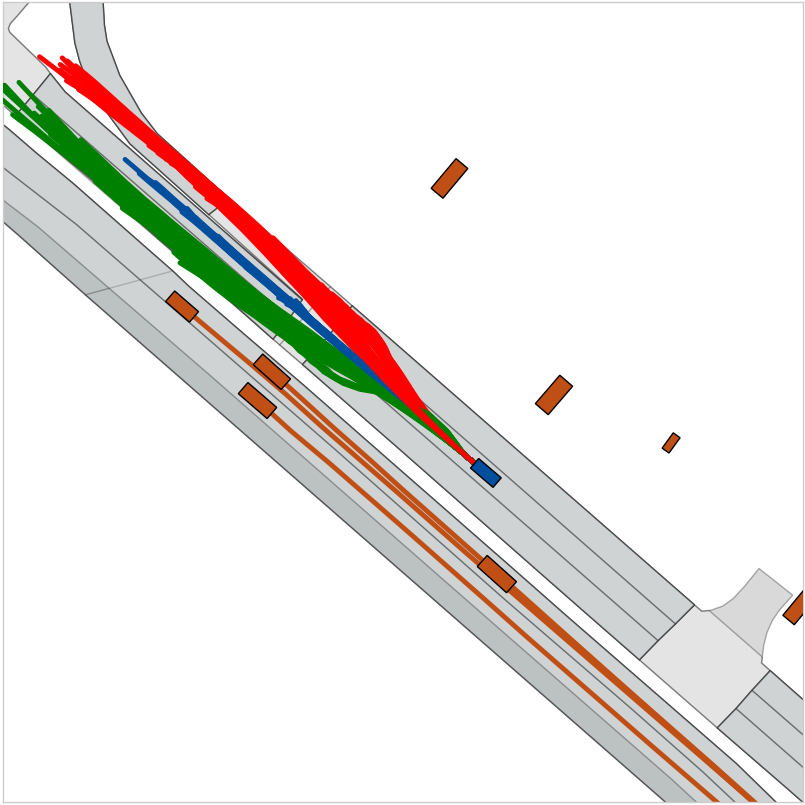}
        \caption{Ours+guidance}
        \label{fig:ours}
    \end{subfigure}
    \begin{subfigure}[b]{0.15\textwidth}
        \centering
        \includegraphics[width=\textwidth]{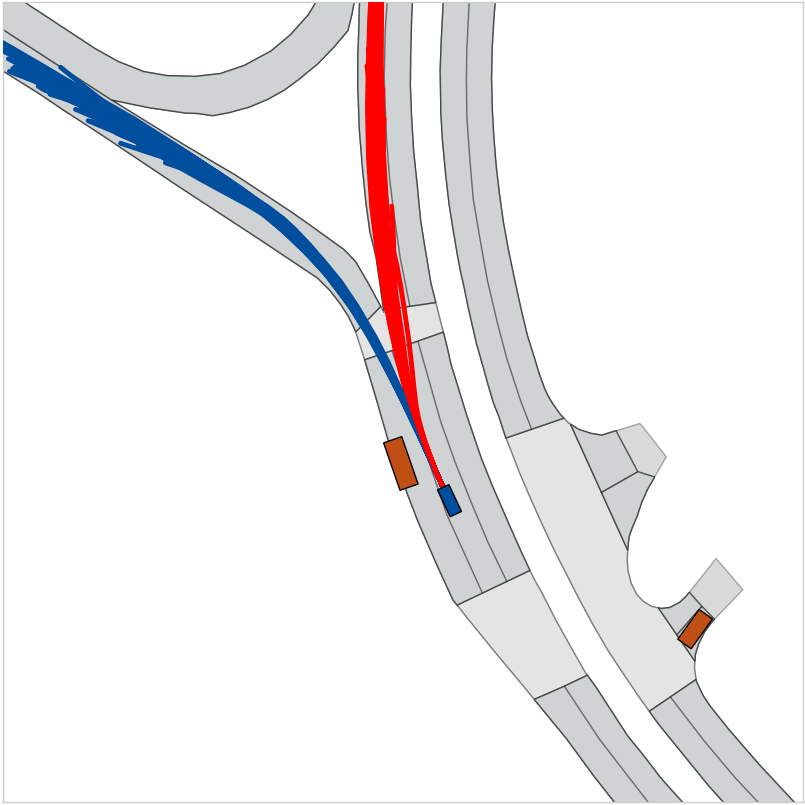}
        \caption{\textit{\trajopt{}}}
        \label{fig:bc1}
    \end{subfigure}
    \begin{subfigure}[b]{0.15\textwidth}
        \centering
        \includegraphics[width=\textwidth]{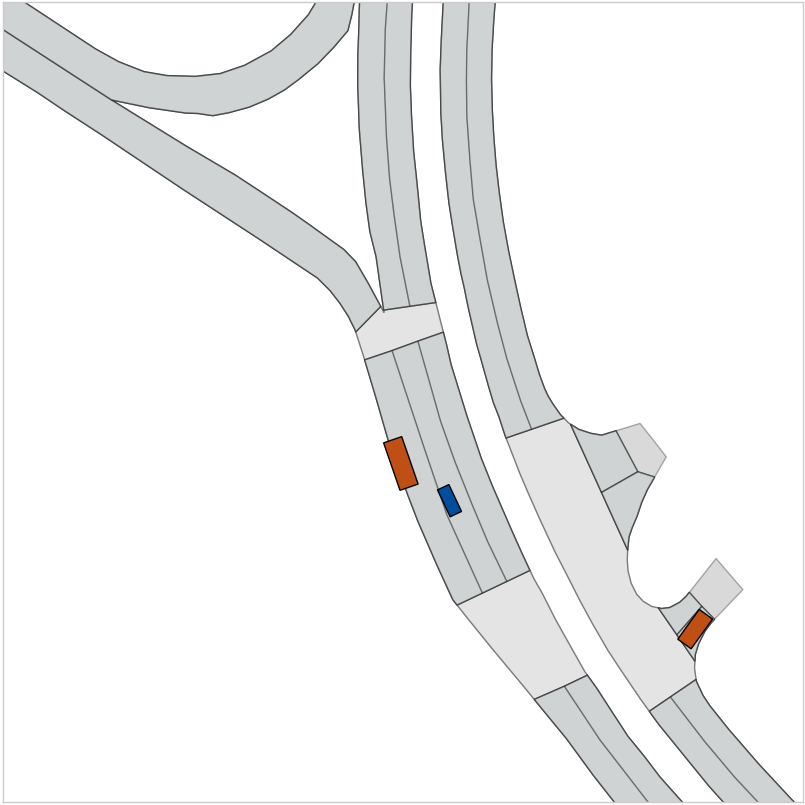}
        \caption{VAE}
        \label{fig:vae01}
    \end{subfigure}
    \begin{subfigure}[b]{0.15\textwidth}
        \centering
        \includegraphics[width=\textwidth]{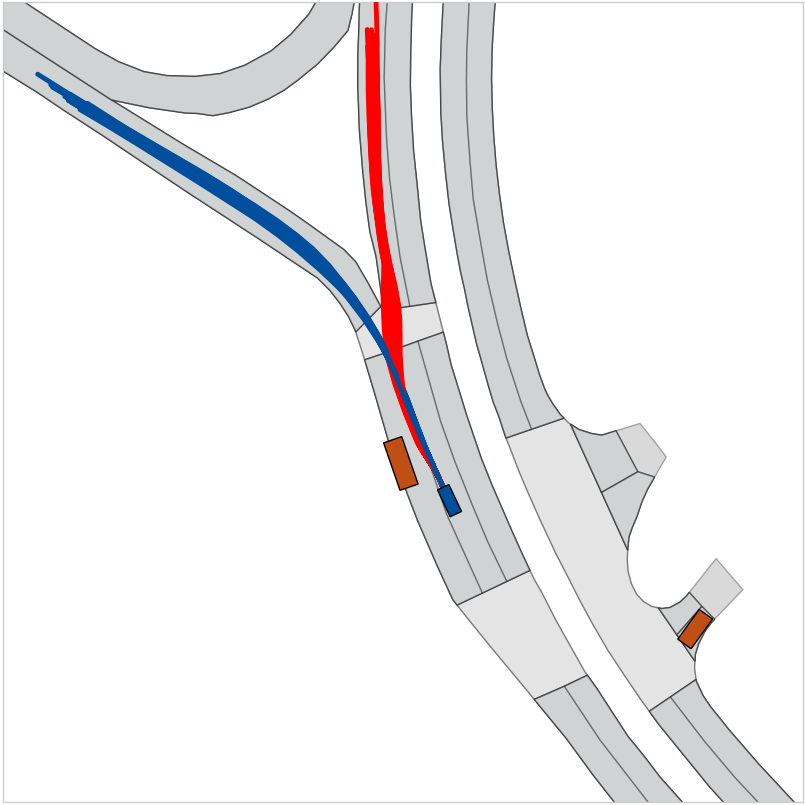}
        \caption{TrafficSim}
        \label{fig:bc-stl1}
    \end{subfigure}
    \begin{subfigure}[b]{0.15\textwidth}
        \centering
        \includegraphics[width=\textwidth]{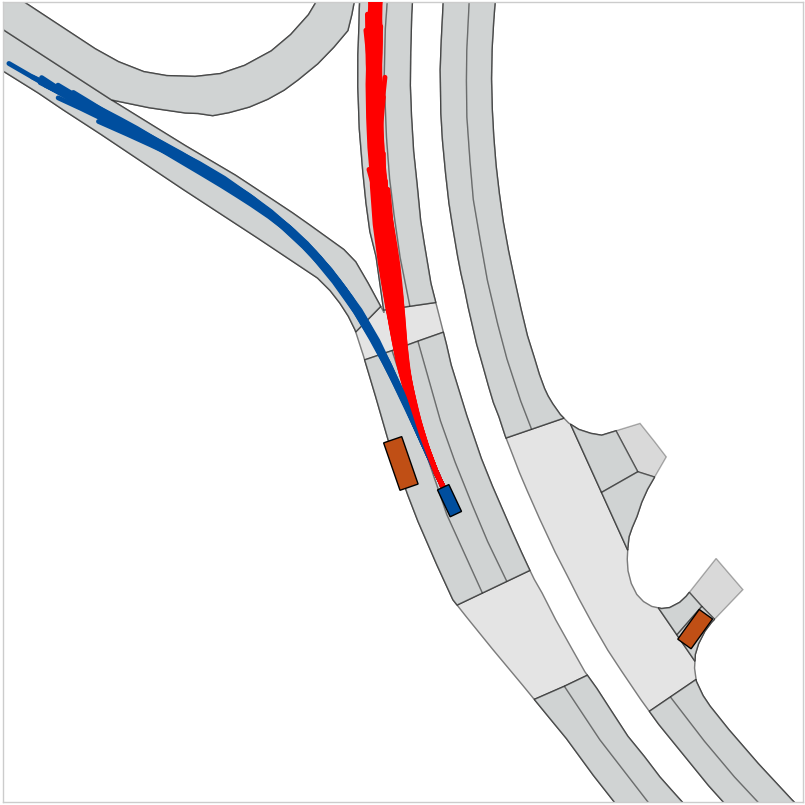}
        \caption{CTG}
        \label{fig:vae1}
    \end{subfigure}
    \begin{subfigure}[b]{0.15\textwidth}
        \centering
        \includegraphics[width=\textwidth]{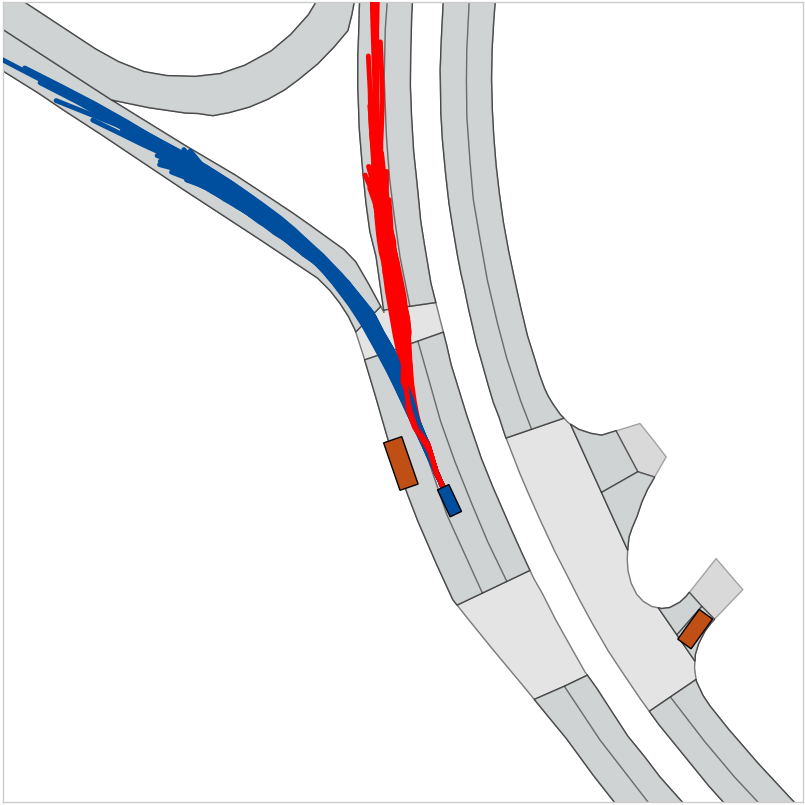}
        \caption{Ours}
        \label{fig:vae-stl1}
    \end{subfigure}
    \begin{subfigure}[b]{0.15\textwidth}
        \centering
        \includegraphics[width=\textwidth]{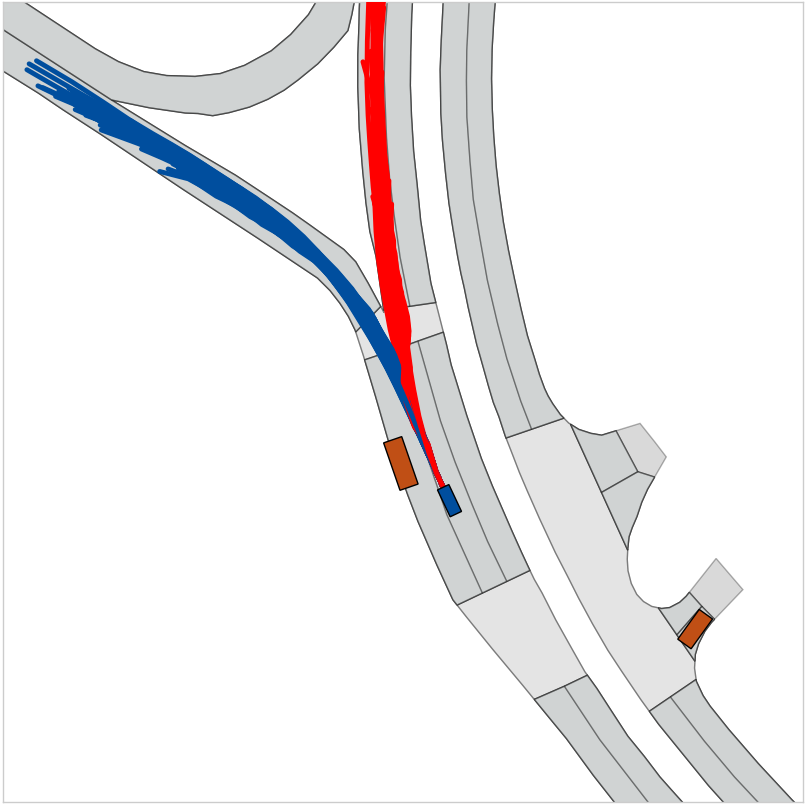}
        \caption{Ours+guidance}
        \label{fig:ours1}
    \end{subfigure}
    \caption{Open-loop visualizations (Green: ``left-lane-change", red: ``right-lane-change" and blue: ``lane-keeping"). Our approach generates the closest to the \textit{\trajopt{}} solution and results in the largest trajectory coverage among all the learning methods.}
    \label{fig:open-loop}
\end{figure*}

We \rbt{first conduct experiments on NuScenes where our method generates the most diverse trajectories quantitatively and visually compared to baselines, reaching the highest STL compliance rate}. In closed-loop test \rbt{we get} the lowest collision rate and out-of-lane rate. Visualization shows how \rbt{varied} STL parameters affect the agent behavior under the same scene, indicating our approach's potential for diverse agent modeling. \rbt{We also consider a human-robot scenario, where we generate the most close-to-oracle distribution. }

\subsection{Implementation details}
\partitle{Dataset} NuScenes~\cite{caesar2020nuscenes} is a large-scale real-world driving dataset that comprises 5.5 hours of driving data from 1000 scenes (from Boston and Singapore). We use the ``trainval" split of the dataset (850 scenes), densely sample from all valid time instants and randomly split the dataset with 70\% for training (11763 samples) and 30\% for validation (5042 samples). \rbt{We developed an annotation tool and it took a student four days to label high-level driving behaviors for the data.} We mainly focus on vehicles and leave the other road participants (pedestrians and cyclists) for future research.

\partitle{Algorithm details} The planning horizon is $T=20$, the duration is $\Delta t=0.5s$, the number of neighbors is $N_n=8$, perception radius is $R=50m$, each lane \rbt{has} $N_p=15$ waypoints, and the control limits are $u_{max}=-u_{min}=(0.5 rad/s, 5.0 m/s^2)^T$. In data generation, the number of samples per scene is $K=64$. \rbt{Similar to~\cite{zhong2023guided}, the diffusion steps is $T_d=100$ and a cosine variance schedule is used.} The networks are FCN with 2 hidden layers, with 256 units for each layer and a ReLU activation for the intermediate layers. For our method, we first generate the augmented dataset, then train the DDPM for 500 epochs using Eq.~\eqref{eq:diffusion-loss}. \rbt{Finally, we freeze the Encoder and DDPM and train the RefineNet for 500 epochs using Eq.~\eqref{eq:diversity-loss}. We use PyTorch with an ADAM~\cite{kingma2014adam} optimizer, a learning rate $3\times 10^{-4}$ and a batch size 128.} The augmentation takes 5 hours, and training takes 8 hours on an RTX4090Ti GPU.

\begin{table*}[!htbp]
\centering
\caption{Closed-loop testing: The highest is shown in \textbf{bold} and the second best is shown in \underline{underline}. ``Ours+guidance" strikes in diversity and rule compliance with the least collision and out-of-lane rate within an acceptable computation budget.}
\begin{tabular}{ccccccc}
\toprule
Methods &  Compliance $\uparrow$ & Valid area $\uparrow$  & Progress $\uparrow$ & Collision $\downarrow$ & Out-of-lane $\downarrow$ & Time (s) $\downarrow$ \\ 
\midrule
VAE & 0.076 & 1.403 & 71.435 & 0.385 & 0.154 & \underline{0.019} \\
DDPM & 0.168 & 5.25\rbt{0} & 83.493 & \underline{0.115} & \textbf{0.000} & 0.031 \\
TrafficSim~\cite{suo2021trafficsim} &0.311 & 3.363 & 71.113 & 0.269 & 0.077 & \textbf{0.018}\\
CTG~\cite{zhong2023guided} & \underline{0.704} & \underline{16.838} & \textbf{92.181} & \underline{0.115} & \textbf{0.000} & 9.280\\
\midrule
Ours & 0.448 & 12.908 & 74.189 & 0.192 & 0.154 & 0.055\\
Ours+guidance & \textbf{0.763} & \textbf{21.577} & \underline{88.638} & \textbf{0.077} & \textbf{0.000} & 0.379\\
\bottomrule
\end{tabular}
\label{tab:closed-loop}
\end{table*}

\subsection{Open-loop evaluation}
\label{sec:exp-open-loop}
\partitle{Baselines} \rbt{The methods are \textbf{\textit{\trajopt{}}}: Trajectory optimization solution (treated as ``Oracle")}; \textbf{VAE}: Variational Auto-encoder;  \textbf{DDPM}: Trained with the loss in Eq.~\eqref{eq:diffusion-loss};  \textbf{TrafficSim}~\cite{suo2021trafficsim}: train a VAE with an extra rule-violation loss; and \textbf{CTG}~\cite{zhong2023guided}: train the DDPM and test with guidance during sampling. And \textbf{Ours}: our method (Sec.~\ref{sec:method-e});  \textbf{Ours+guidance}: with guidance (Sec.~\ref{sec:method-f}). \rbt{For ablations, \textbf{\ournor}: no RefineNet; and \textbf{\ourstl}: uses $\mathcal{L}_{STL}=\text{ReLU}(0.5-\rho)$ to train the RefineNet.} We implement baselines to accommodate for modality and STL rules, and also train VAE and DDPM on the original NuScenes to show gains from our \rbt{augmentation}.


\partitle{Metrics} We evaluate the trajectories by (1) \textbf{Success}: the ratio of the scenes that have at least one trajectory satisfying the STL rules, (2) \textbf{Compliance}: the ratio of generated trajectories satisfying the STL rules (valid trajectories), (3) \textbf{Valid area}: The 2d occupancy area of the valid trajectories averaged over all the scenes, (4) \textbf{Entropy}: \rbt{At each time step, we compute the entropy for the normalized angular velocity and the acceleration from the valid trajectories respectively, and average over all time steps and all scenes}, and (5) \textbf{Time}: measures the trajectory generation time.


\partitle{Quantitative results} As shown in Table~\ref{tab:open-loop}, both VAE and DDPM trained on our augmented dataset achieve a higher diversity (valid area and entropy) than \rbt{them} trained on the original NuScenes data, indicating the value of our augmentation technique to generate diverse demonstrations. VAE and DDPM's low compliance rates (less than 10\%) imply the need to use an advanced model. Compared to advanced baselines TrafficSim~\cite{suo2021trafficsim} and CTG~\cite{zhong2023guided}, \rbt{``Ours"} strikes a sharp rise in the quality and diversity: $32-66\%$ higher for rule compliance, $36-198\%$ larger valid area, and up to $33\%$ increase in entropy. Moreover, ``Ours+guidance" achieves the highest quality and diversity with 1/17X the time used by the best baseline CTG~\footnote{\rbt{``Ours+guidance" is much faster than CTG because we only use guidance at the last five denoising steps, whereas CTG uses guidance at every step.}}.

\partitle{\rbt{Ablation studies}} \rbt{``\ournor" is a bit better than ``DDPM", where the gains result from the ensemble of DDPM outputs from the last five denoising steps. 
With RefineNet and the STL loss used, ``\ourstl" gets a $47-188\%$ increase in the diversity measure compared to ``\ournor". Further using the diversity loss, ``Ours" achieves a $19-22\%$ increase in diversity measure, and ``Ours+guidance" generates the most diverse trajectories but at the cost of 4X longer inference time. 
We can see that adding the RefineNet and using a diversity loss greatly improves the diversity and rule compliance rate (though using the loss $\mathcal{L}_r$ will drop the compliance rate by 3\%.)}
\partitle{\rbt{Visualizations}} In Fig.~\ref{fig:open-loop} we plot all the rule-compliant trajectories (generated by different methods) under specific scenes and color them based on high-level driving modes (red for ``right-lane-change", blue for ``lane-keeping" and green for ``left-lane-change"). ``Ours" and ``Ours+guidance" generate close to \textit{\trajopt{}} distributions, with the largest area coverage among all learning baselines.

\subsection{Closed-loop testing}
\label{sec:exp-closed-loop}

\partitle{Implementation} We select 26 challenging trials in NuScenes dataset (where the ego car needs to avoid cars on the street or to keep track of curvy lanes), and we set the STL parameters \rbt{to the minimum/maximum values in the training data to represent the largest feasible range for parameter selection}. \rbt{We start the simulation from these trials and stop it if} (1) it reaches the max simulation length or (2) collision happens or the ego car drives out-of-lane. \paca{Due to the fast-changing environment, we follow the MPC~\cite{meng2023signal} rather than windowed-policy~\cite{chi2023diffusion,scheikl2024movement} or other mechanism\footnote{\paca{
The works~\cite{chi2023diffusion,scheikl2024movement,reuss2023goal} are mainly for robot manipulation, where the task horizon is long and the environment is relatively static. One challenge in driving scenarios is that the environment can change suddenly in planning (a new neighbor vehicle emerges, lane changes, etc). Thus, the windowed policy might not react to these changes and a mechanism to detect the change and trigger the replanning process is needed. We do not use the goal-conditioned mechanism in~\cite{reuss2023goal} as STL cannot be fully conveyed by a few goal states.
}}~\cite{reuss2023goal}.} At every time step, out of the 64 generated trajectories, we choose the \pac{one} with the highest robustness score and pick its first action to interact with the \rbt{simulator}. We measure: (1) \textbf{Compliance}: the ratio of generated trajectories satisfying the STL rules, (2) \textbf{Valid Area}: The 2d occupancy area of the valid trajectories, (3) \textbf{Progress}: the average driving distance of the ego car, (4) \textbf{Collision}: the ratio of trials ending in collisions, (5) \textbf{Out-of-lane}: the ratio of trials ending in driving out-of-lane, and \pac{(6)} \textbf{Time}: the  runtime at every step.



\partitle{Results} As shown in Table~\ref{tab:closed-loop}, our approach without guidance already achieves high performances compared to VAE, DDPM, and TrafficSim~\cite{suo2021trafficsim}, with slightly high computation time compared to these learning-based baselines - \rbt{the overhead in the runtime mainly owes to the DDPM and STL evaluation}. Given that \rbt{the simulation} $\Delta t=0.5s$, this overhead is still in a reasonable range. With the guidance used, ``Ours+guidance" surpasses all the baselines in quality and diversity metrics (except for CTG's progress), achieving the lowest collision rate and zero out-of-lane. \rbt{Compared to the best baseline, CTG, our runtime is just 1/24X of CTG's. This shows our method's ability to provide diverse and high-quality trajectories with an acceptable time budget in tests.}


\begin{figure}[!htbp]
    \centering
    \begin{subfigure}[b]{0.155\textwidth}
        \centering
        \includegraphics[width=\textwidth]{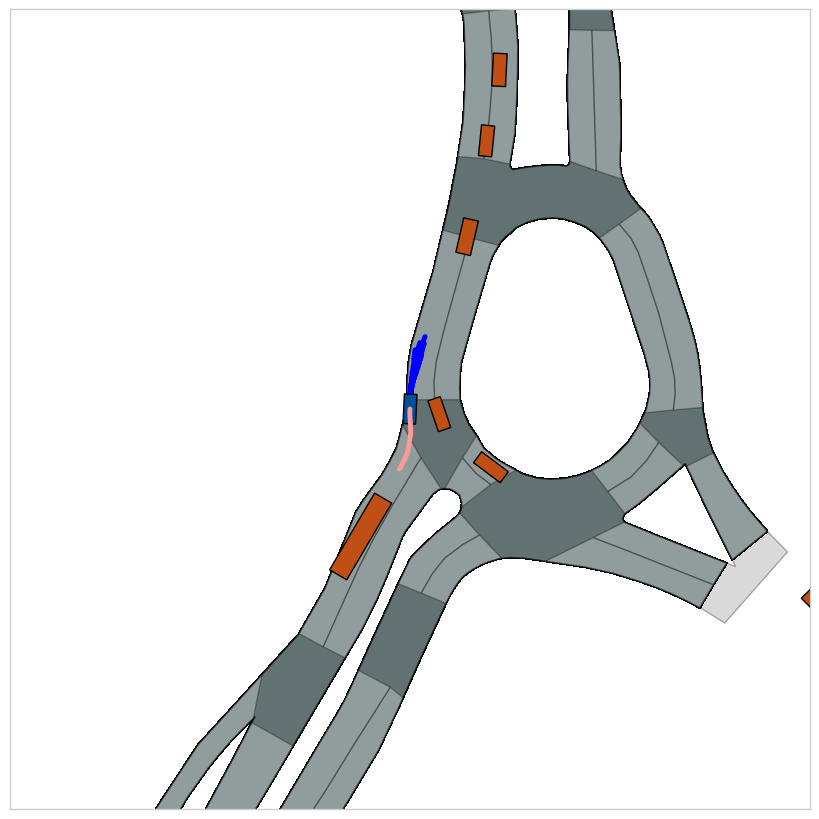}
        \caption{Speed 0 $\sim$ 1m/s}
        \label{fig:testbc1}
    \end{subfigure}
    \begin{subfigure}[b]{0.155\textwidth}
        \centering
        \includegraphics[width=\textwidth]{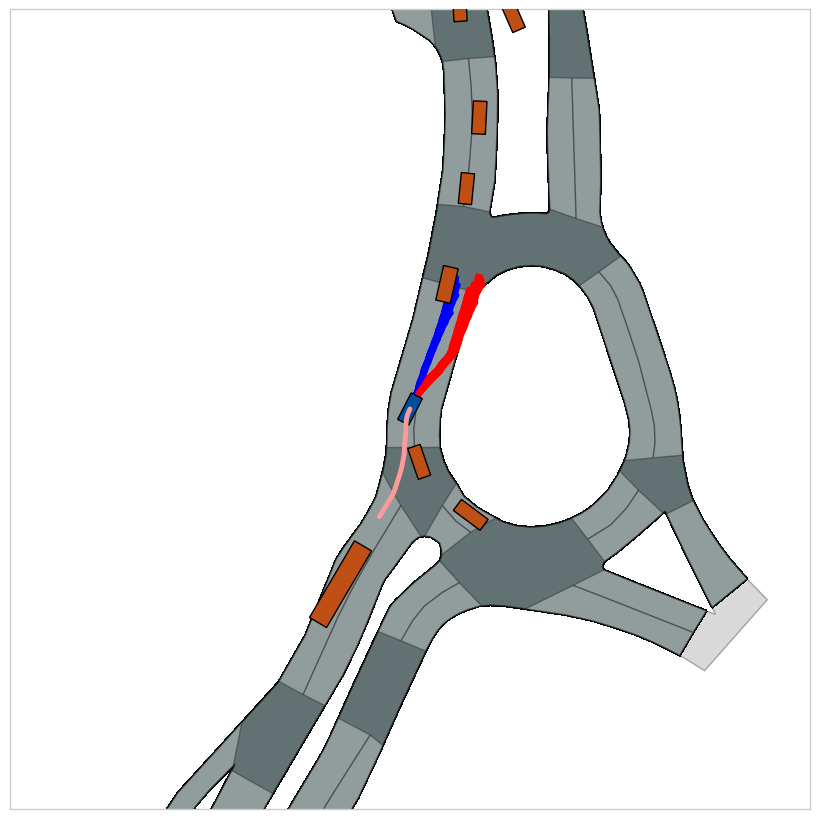}
        \caption{Speed 0 $\sim$ 4m/s}
        \label{fig:testbc2}
    \end{subfigure}
    \begin{subfigure}[b]{0.155\textwidth}
        \centering
        \includegraphics[width=\textwidth]{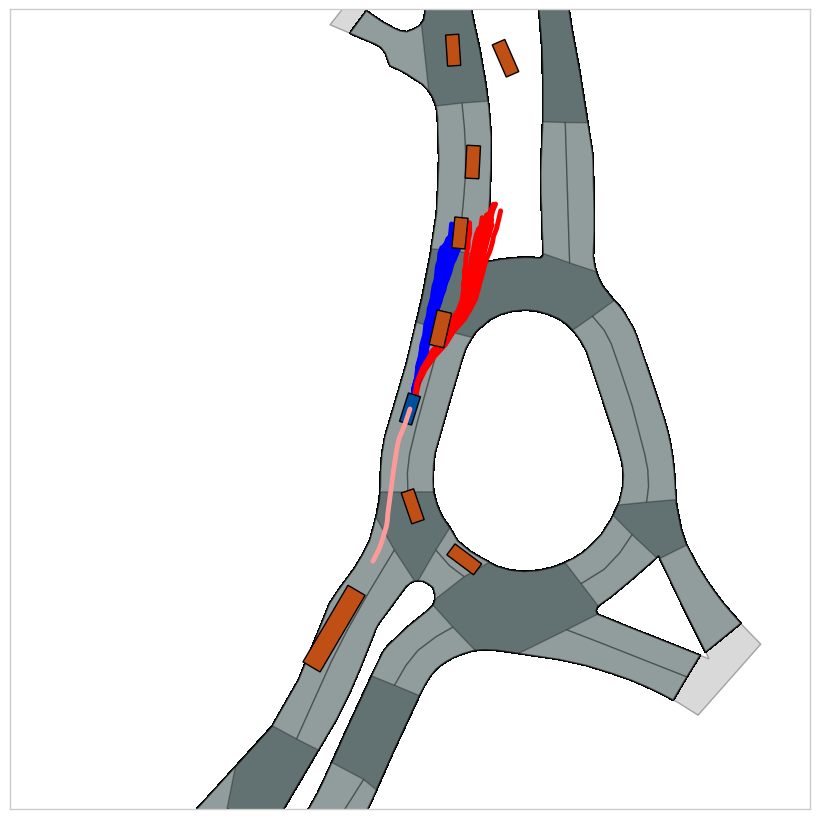}
        \caption{Speed 0 $\sim$ 6m/s}
        \label{fig:testbc3}
    \end{subfigure}
    \caption{Diverse behaviors due to varied STL parameters. When the speed limit is low, the agent waits until all vehicles pass the roundabout. When at the middle-speed limit, the agent joins the queue in the middle but yields to other vehicles at high speed. At the high-speed limit, the car joins the queue and keeps its place as traversing the roundabout.}
    \label{fig:diverse-demo}
\end{figure}

\partitle{Diverse behaviors under different STL parameters}
To show the controllability, we use our method with varied STL parameters to render agent behaviors in a challenging scene shown in Fig.~\ref{fig:diverse-demo}, where the ego car waits to join in a roundabout with dense traffic on the right side. We assign three different maximum speed limits to our network, fix the minimum speed to 0m/s, and plot the scenario at \rbt{t=22}. In Fig.~\ref{fig:diverse-demo}, the history of the ego car is in pink, and planned trajectories are in blue (for lane-keeping) and red (for right-lane-change). When the max speed limit is low (1m/s), the agent waits until all cars finish the roundabout. When at a speed range [0m/s, 4m/s], the agent joins the queue in the middle but yields to other vehicles at high speed. When at a wider interval ([0m/s, 6m/s]), the car joins the middle of the queue and keeps the place as moving in the roundabout. This finding shows we can model different agent characteristics, which is valuable for realistic agent modeling in simulators.

\rbt{\subsection{Case study on diverse human behaviors generation}}
\label{sec:exp-hri}
\rbt{We demonstrate how to generate diverse human behaviors in scenarios~\cite{linard2021formalizing} where the STL rule for the human is to reach the goal while avoiding collision with the incoming robot. 
We choose collision thresholds as STL parameters and calibrate them on 1000 real-world trajectories collected by~\cite{linard2021formalizing}. The dataset is further augmented as in Sec.~\ref{sec:calibration}. 
We train all methods on the augmented dataset for comparison.} 

\begin{table}[!htbp]
\centering
\setlength{\tabcolsep}{2pt}
\caption{\rbt{Open-loop evaluation in human-robot encounter scenarios~\cite{linard2021formalizing}. ``Ours+guidance" reaches the closest diversity to the oracle (\textit{Traj. Opt.}), while using only 1/35X of its time.}}
\rbt{\begin{tabular}{ccccc}
\toprule
Methods & Compliance $\uparrow$ & Area $\uparrow$ &  Entropy $\uparrow$ 
 &Time (s) $\downarrow$  \\ 
\midrule
\textit{\trajopt{}} & \textit{1.000} & \textit{6.114} & \textit{55.553} & \textit{66.125} \\
VAE & 0.420 & 1.894 & 18.600 & \textbf{0.002} \\
DDPM~\cite{ho2020denoising} & 0.262 & 4.673 & 20.715 & 0.023 \\
TrafficSim~\cite{suo2021trafficsim} & \textbf{0.966} & 2.076 & 17.998 & \underline{0.002} \\
CTG~\cite{zhong2023guided} & 0.665 & 4.645 & 39.492 & 16.153 \\
\midrule
Ours & 0.811 & \underline{5.675} & \underline{44.715} & 0.040 \\
Ours+guidance & \underline{0.921} & \textbf{5.773} & \textbf{48.959} & 1.850\\
\bottomrule
\end{tabular}}
\label{tab:rbt-hri-open-loop}
\end{table}

\begin{figure}[!htbp]\captionsetup[subfigure]{font=footnotesize}
    \centering
    \begin{subfigure}[b]{0.075\textwidth}
        \centering
        \includegraphics[width=\textwidth]{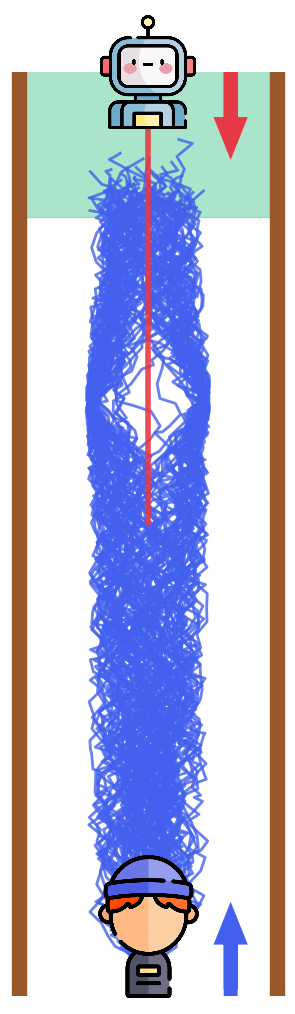}
        \caption{\textit{TrajOpt}}
    \end{subfigure}
    \begin{subfigure}[b]{0.075\textwidth}
        \centering
        \includegraphics[width=\textwidth]{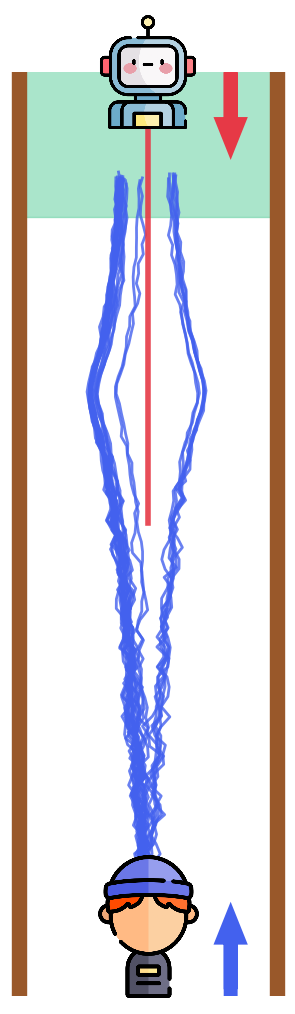}
        \caption{VAE}
        \label{fig:rbt-vae}
    \end{subfigure}
    \begin{subfigure}[b]{0.075\textwidth}
        \centering
        \includegraphics[width=\textwidth]{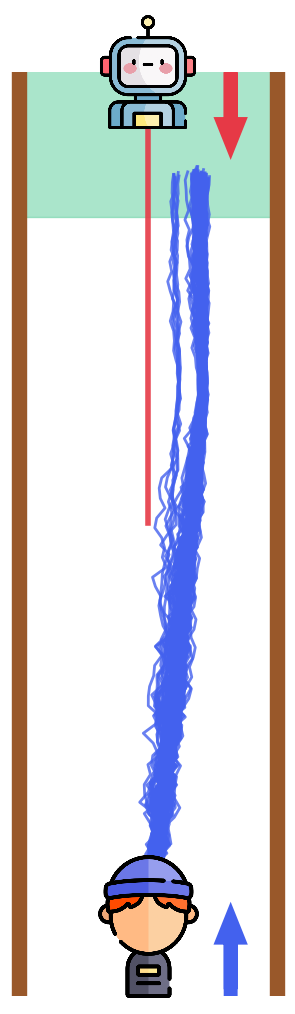}
        \caption{T.S.~\cite{suo2021trafficsim}}
        \label{fig:rbt-trafficsim}
    \end{subfigure}
    \begin{subfigure}[b]{0.075\textwidth}
        \centering
        \includegraphics[width=\textwidth]{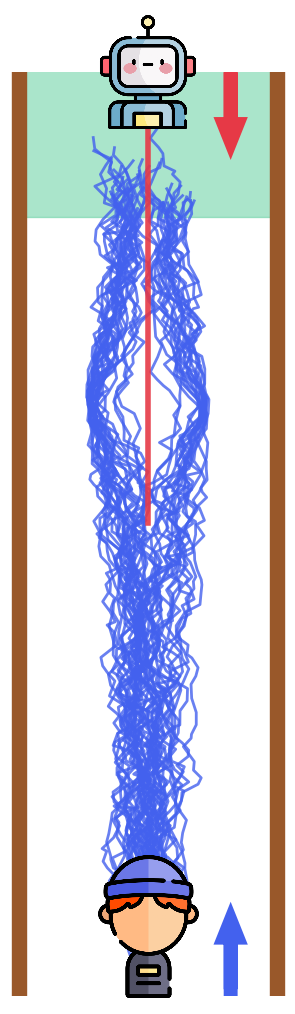}
        \caption{CTG}
    \end{subfigure}
    \begin{subfigure}[b]{0.075\textwidth}
        \centering
        \includegraphics[width=\textwidth]{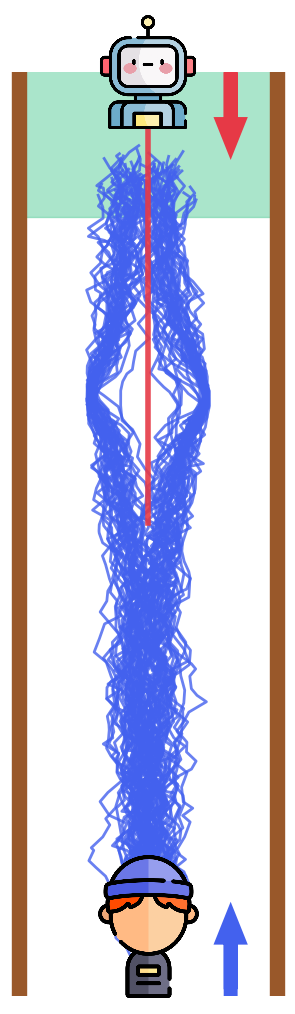}
        \caption{Ours}
    \end{subfigure}
    \begin{subfigure}[b]{0.075\textwidth}
        \centering
        \includegraphics[width=\textwidth]{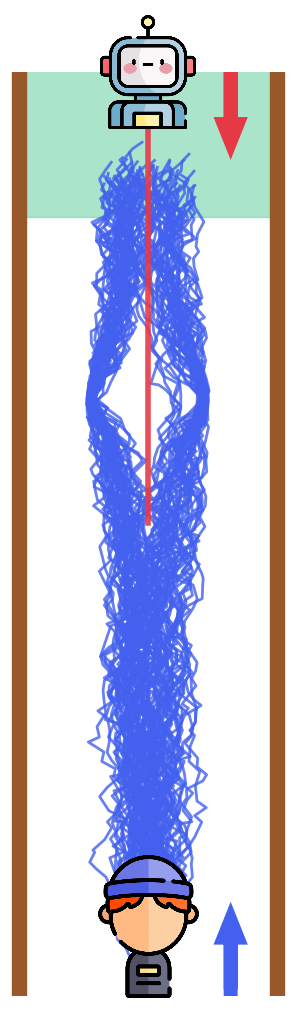}
        \caption{Ours\textsubscript{G}}
    \end{subfigure}
    \caption{\rbt{Valid trajectories for the human-robot encounters. VAE and T.S. (TrafficSim) cannot capture diverse trajectories, whereas Ours\textsubscript{G} (Ours+guidance) are close-to-oracle.}}
    \label{fig:rbt-hri}
\end{figure}

\rbt{In Table~\ref{tab:rbt-hri-open-loop}, ``Ours+guidance" is the highest in ``Area" and ``Entropy", reaching the most close-to-oracle (\textit{Traj. Opt.}) performance with 1/35X of its time. Although TrafficSim~\cite{suo2021trafficsim} reaches the highest compliance rate, its trajectories are less diverse (shown in Fig.~\ref{fig:rbt-trafficsim}). ``Ours+guidance" gets 
$24 \sim 38\%$
improvement over CTG~\cite{zhong2023guided} in rule-compliance rate and diversity, being 7.7X faster in inference speed. Visualizations from Fig.~\ref{fig:rbt-hri} show that the trajectory distribution of Ours\textsubscript{G} (Ours+guidance) is close to the oracle. These results show our advantage in generating diverse rule-compliant policy.}
\section{Conclusions}
\label{sec:conclusion}
We propose a method to learn diverse and rule-compliant agent behavior via data augmentation and \rbt{Diffusion Models}. We model the rules as Signal Temporal Logic (STL), calibrate the STL parameters from the dataset, augment the data using trajectory optimization, and learn the diverse behavior via DDPM and RefineNet. In the NuScenes dataset, we produce the most diverse and rule-compliant trajectories, with 1/17X the runtime used by the second-best baseline~\cite{zhong2023guided}. In closed-loop test, we achieve the highest safety and diversity, and with varied STL parameters we can generate distinct agent behaviors. \rbt{A case study on human-robot scenarios shows we can generate closed-to-oracle trajectories.} The limitations are: \pac{high-level behavior labeling effort;} \rbt{longer runtime than VAE, DDPM, or TrafficSim~\cite{suo2021trafficsim}; and lack of guarantees. 
\pac{Besides, in rare cases if DDPM learns rule-compliance and non-diverse trajectories, our RefineNet cannot improve the diversity.} 
We plan to address those and more complex rules in the future.}

\section*{\pac{Acknowledgement}}
\pac{This work was partly supported by the National Science Foundation (NSF) CAREER Award \#CCF-2238030 and the MIT-Ford Alliance program. Any opinions, findings, conclusions, or recommendations expressed in this publication are those of the authors and don’t reflect the views of the sponsors.}

\bibliographystyle{ieeetr}
\bibliography{z7_references.bib}

\begin{thebibliography}{10}

\bibitem{jager2002need}
W.~Jager and M.~Janssen, ``The need for and development of behaviourally realistic agents,'' in {\em International Workshop on Multi-Agent Systems and Agent-Based Simulation}, pp.~36--49, Springer, 2002.

\bibitem{suo2021trafficsim}
S.~Suo, S.~Regalado, S.~Casas, and R.~Urtasun, ``Trafficsim: Learning to simulate realistic multi-agent behaviors,'' in {\em Proceedings of the IEEE/CVF Conference on Computer Vision and Pattern Recognition}, pp.~10400--10409, 2021.

\bibitem{dosovitskiy2017carla}
A.~Dosovitskiy, G.~Ros, F.~Codevilla, A.~Lopez, and V.~Koltun, ``Carla: An open urban driving simulator,'' in {\em Conference on robot learning}, pp.~1--16, PMLR, 2017.

\bibitem{lopez2018microscopic}
P.~A. Lopez, M.~Behrisch, L.~Bieker-Walz, J.~Erdmann, Y.-P. Fl{\"o}tter{\"o}d, R.~Hilbrich, L.~L{\"u}cken, J.~Rummel, P.~Wagner, and E.~Wie{\ss}ner, ``Microscopic traffic simulation using sumo,'' in {\em 2018 21st international conference on intelligent transportation systems (ITSC)}, 2018.

\bibitem{kesting2010enhanced}
A.~KESTING and M.~T.~D. HELBING, ``Enhanced intelligent driver model to access the impact of driving strategies on traffic capacity,'' {\em Phil. Trans. R. Soc. A}, vol.~368, pp.~4585--4605, 2010.

\bibitem{kesting2007general}
A.~Kesting, M.~Treiber, and D.~Helbing, ``General lane-changing model mobil for car-following models,'' {\em Transportation Research Record}, vol.~1999, no.~1, pp.~86--94, 2007.

\bibitem{salzmann2020trajectron++}
T.~Salzmann, B.~Ivanovic, P.~Chakravarty, and M.~Pavone, ``Trajectron++: Dynamically-feasible trajectory forecasting with heterogeneous data,'' in {\em European Conference on Computer Vision}, 2020.

\bibitem{bhattacharyya2022modeling}
R.~Bhattacharyya, B.~Wulfe, D.~J. Phillips, A.~Kuefler, J.~Morton, R.~Senanayake, and M.~J. Kochenderfer, ``Modeling human driving behavior through generative adversarial imitation learning,'' {\em IEEE Transactions on Intelligent Transportation Systems}, 2022.

\bibitem{liang2020garden}
J.~Liang, L.~Jiang, K.~Murphy, T.~Yu, and A.~Hauptmann, ``The garden of forking paths: Towards multi-future trajectory prediction,'' in {\em Proceedings of the IEEE/CVF Conference on Computer Vision and Pattern Recognition}, pp.~10508--10518, 2020.

\bibitem{donze2010robust}
A.~Donz{\'e} and O.~Maler, ``Robust satisfaction of temporal logic over real-valued signals,'' {\em Formal Modeling and Analysis of Timed Systems}, p.~92, 2010.

\bibitem{donze2013efficient}
A.~Donz{\'e}, T.~Ferrere, and O.~Maler, ``Efficient robust monitoring for stl,'' in {\em Computer Aided Verification: 25th International Conference, CAV 2013}, pp.~264--279, Springer, 2013.

\bibitem{sahin2020autonomous}
Y.~E. Sahin, R.~Quirynen, and S.~Di~Cairano, ``Autonomous vehicle decision-making and monitoring based on signal temporal logic and mixed-integer programming,'' in {\em 2020 American Control Conference (ACC)}, pp.~454--459, IEEE, 2020.

\bibitem{dawson2022robust}
C.~Dawson and C.~Fan, ``Robust counterexample-guided optimization for planning from differentiable temporal logic,'' in {\em 2022 IEEE/RSJ International Conference on Intelligent Robots and Systems (IROS)}, pp.~7205--7212, IEEE, 2022.

\bibitem{leung2023backpropagation}
K.~Leung, N.~Ar{\'e}chiga, and M.~Pavone, ``Backpropagation through signal temporal logic specifications: Infusing logical structure into gradient-based methods,'' {\em The International Journal of Robotics Research}, vol.~42, no.~6, pp.~356--370, 2023.

\bibitem{meng2023signal}
Y.~Meng and C.~Fan, ``Signal temporal logic neural predictive control,'' {\em IEEE Robotics and Automation Letters}, 2023.

\bibitem{li2017reinforcement}
X.~Li, C.-I. Vasile, and C.~Belta, ``Reinforcement learning with temporal logic rewards,'' in {\em 2017 IEEE/RSJ International Conference on Intelligent Robots and Systems (IROS)}, pp.~3834--3839, IEEE, 2017.

\bibitem{ho2020denoising}
J.~Ho, A.~Jain, and P.~Abbeel, ``Denoising diffusion probabilistic models,'' {\em Advances in neural information processing systems}, 2020.

\bibitem{zhong2023guided}
Z.~Zhong, D.~Rempe, D.~Xu, Y.~Chen, S.~Veer, T.~Che, B.~Ray, and M.~Pavone, ``Guided conditional diffusion for controllable traffic simulation,'' in {\em 2023 IEEE International Conference on Robotics and Automation (ICRA)}, pp.~3560--3566, IEEE, 2023.

\bibitem{chi2023diffusion}
C.~Chi, S.~Feng, Y.~Du, Z.~Xu, E.~Cousineau, B.~Burchfiel, and S.~Song, ``Diffusion policy: Visuomotor policy learning via action diffusion,'' {\em arXiv preprint arXiv:2303.04137}, 2023.

\bibitem{caesar2020nuscenes}
H.~Caesar, V.~Bankiti, A.~H. Lang, S.~Vora, V.~E. Liong, Q.~Xu, A.~Krishnan, Y.~Pan, G.~Baldan, and O.~Beijbom, ``nuscenes: A multimodal dataset for autonomous driving,'' in {\em Proceedings of the IEEE/CVF conference on computer vision and pattern recognition}, pp.~11621--11631, 2020.

\bibitem{leon2021review}
F.~Leon and M.~Gavrilescu, ``A review of tracking and trajectory prediction methods for autonomous driving,'' {\em Mathematics}, vol.~9, no.~6, p.~660, 2021.

\bibitem{ammoun2009real}
S.~Ammoun and F.~Nashashibi, ``Real time trajectory prediction for collision risk estimation between vehicles,'' in {\em 2009 IEEE 5Th international conference on intelligent computer communication and processing}, pp.~417--422, IEEE, 2009.

\bibitem{joseph2011bayesian}
J.~Joseph, F.~Doshi-Velez, A.~S. Huang, and N.~Roy, ``A bayesian nonparametric approach to modeling motion patterns,'' {\em Autonomous Robots}, vol.~31, pp.~383--400, 2011.

\bibitem{ly2020learning}
A.~O. Ly and M.~Akhloufi, ``Learning to drive by imitation: An overview of deep behavior cloning methods,'' {\em IEEE Transactions on Intelligent Vehicles}, vol.~6, no.~2, pp.~195--209, 2020.

\bibitem{sun2020scalability}
P.~Sun, H.~Kretzschmar, X.~Dotiwalla, A.~Chouard, V.~Patnaik, P.~Tsui, J.~Guo, Y.~Zhou, Y.~Chai, B.~Caine, {\em et~al.}, ``Scalability in perception for autonomous driving: Waymo open dataset,'' {\em arXiv e-prints}, 2019.

\bibitem{gao2020vectornet}
J.~Gao, C.~Sun, H.~Zhao, Y.~Shen, D.~Anguelov, C.~Li, and C.~Schmid, ``Vectornet: Encoding hd maps and agent dynamics from vectorized representation,'' in {\em Proceedings of the IEEE/CVF Conference on Computer Vision and Pattern Recognition}, pp.~11525--11533, 2020.

\bibitem{gupta2018social}
A.~Gupta, J.~Johnson, L.~Fei-Fei, S.~Savarese, and A.~Alahi, ``Social gan: Socially acceptable trajectories with generative adversarial networks,'' in {\em Proceedings of the IEEE conference on computer vision and pattern recognition}, pp.~2255--2264, 2018.

\bibitem{phan2020covernet}
T.~Phan-Minh, E.~C. Grigore, F.~A. Boulton, O.~Beijbom, and E.~M. Wolff, ``Covernet: Multimodal behavior prediction using trajectory sets,'' in {\em Proceedings of the IEEE/CVF conference on computer vision and pattern recognition}, pp.~14074--14083, 2020.

\bibitem{gilles2021thomas}
T.~Gilles, S.~Sabatini, D.~Tsishkou, B.~Stanciulescu, and F.~Moutarde, ``Thomas: Trajectory heatmap output with learned multi-agent sampling,'' {\em arXiv preprint arXiv:2110.06607}, 2021.

\bibitem{jiang2023motiondiffuser}
C.~Jiang, A.~Cornman, C.~Park, B.~Sapp, Y.~Zhou, D.~Anguelov, {\em et~al.}, ``Motiondiffuser: Controllable multi-agent motion prediction using diffusion,'' in {\em Proceedings of the IEEE/CVF Conference on Computer Vision and Pattern Recognition}, pp.~9644--9653, 2023.

\bibitem{reuss2023goal}
M.~Reuss, M.~Li, X.~Jia, and R.~Lioutikov, ``Goal-conditioned imitation learning using score-based diffusion policies,'' {\em arXiv preprint arXiv:2304.02532}, 2023.

\bibitem{scheikl2024movement}
P.~M. Scheikl, N.~Schreiber, C.~Haas, N.~Freymuth, G.~Neumann, R.~Lioutikov, and F.~Mathis-Ullrich, ``Movement primitive diffusion: Learning gentle robotic manipulation of deformable objects,'' {\em IEEE Robotics and Automation Letters}, 2024.

\bibitem{yuan2019diverse}
Y.~Yuan and K.~Kitani, ``Diverse trajectory forecasting with determinantal point processes,'' {\em arXiv preprint arXiv:1907.04967}, 2019.

\bibitem{xu2024controllable}
Y.~Xu, H.~Cheng, and M.~Sester, ``Controllable diverse sampling for diffusion based motion behavior forecasting,'' {\em arXiv preprint arXiv:2402.03981}, 2024.

\bibitem{kim2022diverse}
S.~Kim, H.~Jeon, J.~W. Choi, and D.~Kum, ``Diverse multiple trajectory prediction using a two-stage prediction network trained with lane loss,'' {\em IEEE Robotics and Automation Letters}, pp.~2038--2045, 2022.

\bibitem{stoll2023scaling}
M.~Stoll, M.~Mazzola, M.~Dolgov, J.~Mathes, and N.~M{\"o}ser, ``Scaling planning for automated driving using simplistic synthetic data,'' {\em arXiv preprint arXiv:2305.18942}, 2023.

\bibitem{meng2021reactive}
Y.~Meng, Z.~Qin, and C.~Fan, ``Reactive and safe road user simulations using neural barrier certificates,'' in {\em 2021 IEEE/RSJ International Conference on Intelligent Robots and Systems (IROS)}, IEEE, 2021.

\bibitem{casas2020importance}
S.~Casas, C.~Gulino, S.~Suo, and R.~Urtasun, ``The importance of prior knowledge in precise multimodal prediction,'' in {\em 2020 IEEE/RSJ International Conference on Intelligent Robots and Systems (IROS)}, pp.~2295--2302, IEEE, 2020.

\bibitem{bhattacharyya2019simulating}
R.~P. Bhattacharyya, D.~J. Phillips, C.~Liu, J.~K. Gupta, K.~Driggs-Campbell, and M.~J. Kochenderfer, ``Simulating emergent properties of human driving behavior using multi-agent reward augmented imitation learning,'' in {\em 2019 International Conference on Robotics and Automation (ICRA)}, pp.~789--795, IEEE, 2019.

\bibitem{zhong2023language}
Z.~Zhong, D.~Rempe, Y.~Chen, B.~Ivanovic, Y.~Cao, D.~Xu, M.~Pavone, and B.~Ray, ``Language-guided traffic simulation via scene-level diffusion,'' {\em arXiv preprint arXiv:2306.06344}, 2023.

\bibitem{maierhofer2022formalization}
S.~Maierhofer, P.~Moosbrugger, and M.~Althoff, ``Formalization of intersection traffic rules in temporal logic,'' in {\em 2022 IEEE Intelligent Vehicles Symposium (IV)}, pp.~1135--1144, IEEE, 2022.

\bibitem{maler2004monitoring}
O.~Maler and D.~Nickovic, ``Monitoring temporal properties of continuous signals,'' {\em Formal Techniques, ModellingandAnalysis of Timed and Fault-Tolerant Systems}, p.~152, 2004.

\bibitem{pant2017smooth}
Y.~V. Pant, H.~Abbas, and R.~Mangharam, ``Smooth operator: Control using the smooth robustness of temporal logic,'' in {\em 2017 IEEE Conference on Control Technology and Applications}, IEEE, 2017.

\bibitem{qi2017pointnet}
C.~R. Qi, H.~Su, K.~Mo, and L.~J. Guibas, ``Pointnet: Deep learning on point sets for 3d classification and segmentation,'' in {\em Proceedings of the IEEE conference on computer vision and pattern recognition}, pp.~652--660, 2017.

\bibitem{kingma2014adam}
D.~P. Kingma and J.~Ba, ``Adam: A method for stochastic optimization,'' {\em arXiv preprint arXiv:1412.6980}, 2014.

\bibitem{linard2021formalizing}
A.~Linard, I.~Torre, A.~Steen, I.~Leite, and J.~Tumova, ``Formalizing trajectories in human-robot encounters via probabilistic stl inference,'' in {\em 2021 IEEE/RSJ International Conference on Intelligent Robots and Systems (IROS)}, pp.~9857--9862, IEEE, 2021.

\end{thebibliography}
\end{document}